\PassOptionsToPackage{table}{xcolor}
\documentclass[10pt,twocolumn,letterpaper]{article}

\usepackage[pagenumbers]{wacv} % To force page numbers, e.g. for an arXiv version

\usepackage[accsupp]{axessibility}
\usepackage{amsmath, amssymb, amsthm}
\usepackage{bm}
\usepackage{comment}
\usepackage{booktabs}      % \toprule, \midrule, \bottomrule, \cmidrule
\usepackage{graphicx}      % \resizebox
\usepackage{array}         % better column types
\usepackage{textcomp}
\usepackage{siunitx}
\usepackage{multirow}

\definecolor{best}{RGB}{255,140,140}
\definecolor{second}{RGB}{255,190,120}
\definecolor{third}{RGB}{255,235,140}
\newcommand{\cbest}[1]{\cellcolor{best}#1}
\newcommand{\csecond}[1]{\cellcolor{second}#1}
\newcommand{\cthird}[1]{\cellcolor{third}#1}

\usepackage{tikz}
\usetikzlibrary{arrows.meta,calc,positioning,fit,backgrounds}

\newcommand{\SO}{\mathrm{SO}(3)}
\newcommand{\so}{\mathfrak{so}(3)}
\newcommand{\R}{\mathbb{R}}
\newcommand{\Exp}{\mathrm{Exp}}
\newcommand{\Log}{\mathrm{Log}}

\definecolor{wacvblue}{rgb}{0.21,0.49,0.74}
\usepackage[pagebackref,breaklinks,colorlinks,allcolors=wacvblue]{hyperref}

\def\wacvPaperID{89} % *** Enter the WACV Paper ID here
\def\confName{WACV}
\def\confYear{2027}

\title{PoseRieFlow: Riemannian Flow Matching for Whole-Body Text-to-Pose Generation}

\author{Jan Spindler \qquad Bardia Ardakanian \qquad Patrick Stotko \qquad Reinhard Klein\\
University of Bonn\\
{\tt\small jspindle@cs.uni-bonn.de \quad ardakanian@uni-bonn.de \quad \{stotko,rk\}@cs.uni-bonn.de}
}

\begin{document}
\maketitle
\begin{abstract}
We present PoseRieFlow, a compact flow-matching framework for text-conditioned generation of whole-body 3D human poses. Prior text-to-pose methods either neglect articulated hands or model rotations in Euclidean space despite their non-Euclidean geometry. PoseRieFlow instead formulates generation directly on the product manifold of SMPL-H joint rotations, learning a text-conditioned vector field that transports samples from an isotropic $\mathrm{SO}(3)$ Gaussian prior to the data distribution along Riemannian geodesics. A part-tokenized Diffusion Transformer aggregates the 52 joints into eight body-part tokens, allocating a dedicated token and nonlinear projector to each hand, and restores per-joint resolution through part-specific decoders, shortening the sequence while improving kinematic fidelity over per-joint tokenization. With only 21.6M trainable parameters, PoseRieFlow supports expressive SMPL-H generation with detailed hand articulation at a fraction of the scale of recent LLM-based generators. On PoseScript, it attains the lowest whole-body kinematic error among all hand-producing methods, with consistent kinematic gains from the Riemannian formulation over its Euclidean counterpart at every practical inference budget, while distributional and retrieval metrics remain comparable between them.
\end{abstract}
    
\section{Introduction}
\label{sec:intro}

Generating 3D human poses from natural language is a fundamental problem spanning vision, graphics, and embodied AI. Recent progress in text-conditioned generative modeling has advanced text-to-motion synthesis through diffusion and flow-based approaches~\cite{tevet2023human, gao2025motionflux, hymotion2025}. In contrast, text-to-pose generation~\cite{delmas2024posescript, feng2024chatpose, li2025unipose} remains comparatively underexplored despite posing a distinct challenge: unlike motion generation, which benefits from strong temporal continuity priors, single-frame pose generation must model a highly multimodal distribution over articulated 3D configurations from language alone. This challenge becomes harder when extending from coarse body posture to whole-body articulation with expressive hand poses.

Current text-to-pose methods exhibit two limitations. First, most approaches model joint rotations in Euclidean parameterizations such as axis-angle or continuous 6D rotations~\cite{zhou20196d}, and optimize Euclidean regression objectives. However, articulated pose lies on a product manifold of rotation groups rather than a flat vector space. As a result, linear interpolation and Euclidean velocity regression produce optimization targets that are not aligned with the geodesic structure of rotations, particularly for large joint rotations and high-dimensional whole-body configurations. Recent work on geometry-aware articulated modeling~\cite{he2024nrdf, miao2026riemannian} suggests that respecting manifold structure can improve generative learning, but these ideas have not yet been explored in text-conditioned single-frame pose generation.

Second, existing text-to-pose systems largely ignore detailed hand articulation. Most prior methods operate on body-only skeletons with 22 joints, even when datasets are stored in richer SMPL-H or SMPL-X formats~\cite{looper2015smpl, romero2017mano, pavlakos2019smplx, delmas2024posescript, delmas2023posefix}. Recent LLM-based generators~\cite{feng2024chatpose, li2025unipose} similarly omit finger degrees of freedom. Yet hand articulation carries a substantial fraction of pose semantics in natural language descriptions, including gestures such as pointing, grasping, waving, or open-palm interactions. Modeling full SMPL-H articulation therefore requires learning coherent dependencies across 52 coupled rotations spanning body posture and dexterous hand structure. At the same time, naively scaling transformer architectures to whole-body pose increases sequence length and attention cost, motivating architectures that better exploit anatomical structure as an inductive bias.

We address these challenges with \emph{PoseRieFlow}, a Riemannian flow-matching framework for text-conditioned whole-body 3D pose generation. PoseRieFlow formulates generation directly on the product manifold $\SO^{52}$ of SMPL-H joint rotations, learning a text-conditioned vector field that transports samples from an isotropic Riemannian Gaussian prior to the data distribution along geodesic paths. Instead of regressing Euclidean rotation updates, the model predicts tangent vectors in the local Lie algebra of each joint, yielding training signals aligned with the geometry of articulated rotations. To efficiently model high-dimensional whole-body structure, we introduce a part-tokenized Diffusion Transformer that aggregates joints into semantic body-part tokens while preserving per-joint outputs through lightweight decoders. PoseRieFlow thereby bridges geometry-aware generative modeling and expressive whole-body text-to-pose synthesis.

Our contributions are as follows:
\begin{itemize}
    \item We introduce \emph{PoseRieFlow}, a flow-matching model for text-conditioned whole-body 3D pose generation on the product manifold of 52 SMPL-H joint rotations. The model learns a text-conditioned vector field between an isotropic $\SO$ Gaussian prior and the data distribution along Riemannian geodesics, yielding lower rotational and joint-position error than Euclidean flow matching at every practical inference budget.

    \item We bring part-based tokenization~\cite{wang2023iip,dou2023tore,wang2024gtpt} from discriminative pose models to a generative Diffusion Transformer. As $30$ of the $52$ SMPL-H joints are fingers, we give each hand its own token with a nonlinear projector, and recover per-joint resolution through part-specific decoders. This shortens the attention sequence from 52 to 8 tokens while improving kinematic fidelity over per-joint tokenization at matched data.

    \item We demonstrate that geometry-aware whole-body pose generation can be achieved with a lightweight task-specific architecture. PoseRieFlow contains only 21.6M trainable parameters, whereas recent text-to-pose systems have billions, while additionally supporting full SMPL-H generation with articulated hands.
\end{itemize}

\section{Related Work}
\label{sec:related}

\paragraph{Text-conditioned single-frame pose generation.}
Text-to-pose generation maps natural language to static 3D human poses, a building block for motion synthesis and character control. PoseScript~\cite{delmas2024posescript} introduced the canonical benchmark, pairing 100k SMPL-H poses from AMASS~\cite{AMASS} with human-written and automatic captions, and proposed a conditional VAE. Extensions include PoseFix~\cite{delmas2023posefix} for text-guided editing and PoseEmbroider~\cite{delmas2024poseembroider} for multimodal embeddings. More recent systems generate pose tokens autoregressively with billion-parameter language backbones, including ChatPose~\cite{feng2024chatpose}, PoseLLaVA~\cite{posellava2025}, UniPose~\cite{li2025unipose}, and CoT-Pose~\cite{cha2025cotpose}. These methods focus on body-only generation and do not model articulated hands, limiting the pose semantics they can capture.

\paragraph{Geometry-aware generative modeling on manifolds.}
Human pose naturally lives on a product manifold of joint rotations. NRDF~\cite{he2024nrdf} learned neural Riemannian distance fields on the quaternion manifold as an articulated pose prior, extended to motion by Geometric Neural Distance Fields~\cite{yu2025geometric}. Flow matching~\cite{lipman2023flow} learns continuous vector fields between distributions without simulation and was generalized to Riemannian manifolds via geodesic interpolation~\cite{chen2024flow}. On Lie groups, denoising diffusion and flow matching over $\mathrm{SO}(3)$ and $\mathrm{SE}(3)$ have been applied to rotational alignment~\cite{leach2022so3}, protein backbone generation~\cite{bose2024se3, yim2023se3}, and robot motion~\cite{braun2024riemannianfm}. In human motion, Motion Flow Matching~\cite{hu2023motion}, MotionFLUX~\cite{gao2025motionflux}, and HY-Motion~1.0~\cite{hymotion2025} scale Euclidean flow matching, and, closest to our work, RMG~\cite{miao2026riemannian} applied Riemannian flow matching to text-to-motion on HumanML3D~\cite{guo2022humanml3d}. Rectified-flow variants straighten transport paths for few-step sampling~\cite{liu2023rectified, esser2024sd3}, an axis orthogonal to the choice of geometry, which we study via geodesic rectification in the supplementary material. We differ in three respects: we target single-frame generation, where no temporal prior constrains the output, we model full SMPL-H articulation including fingers, and we provide a controlled Euclidean vs.\ Riemannian comparison at matched architecture, data, and compute, isolating the effect of geometry-aware learning.

\paragraph{SMPL-H and hand-aware pose generation.}
Hand articulation carries much of the semantics in natural language, such as pointing and grasping, yet most text-to-pose methods ignore finger degrees of freedom. HanDiffuser~\cite{narasimhaswamy2024handiffuser} targets hand-realistic image synthesis rather than pose benchmarks. BOTH2Hands~\cite{zhang2024both2hands} synthesizes two-hand motion from body dynamics and text, and MotionCraft~\cite{bian2025motioncraft} generates SMPL-X whole-body motion with plug-and-play multimodal controls. Both target motion with specialised control signals, whereas we study single-frame text-to-pose.

\paragraph{Part-based and grouped-token architectures.}
Grouping joints into semantic parts is a long-standing inductive bias~\cite{du2015hierarchical,jain2016srnn,thakkar2018pbgcn}, realised in transformers either as one token per joint~\cite{li2021tokenpose,aksan2021sttransformer,zhong2023attt2m} or one token per part. IIP-Transformer~\cite{wang2023iip} encodes 25 joints as 5 tokens, cutting attention cost, and TORE~\cite{dou2023tore} and GTPT~\cite{wang2024gtpt} confirm that token reduction pays off for mesh recovery and whole-body keypoints, with the largest gains on keypoint-dense regions such as the hands~\cite{zeng2022tcformer,wang2024gtpt}. Part-level representations likewise underpin recent motion--language retrieval models, which form transformer patches from body-part joint groups~\cite{yu2024motionpatches} or retrieve reference motions by part-level features~\cite{yu2025remogpt}. In generative human modelling, by contrast, parts have not been placed on the sequence axis of a single transformer: prior systems either give each part its own codebook or generator network~\cite{zou2024parco,siyao2022bailando,yi2023talkshow,liu2024emage,lu2024humantomato}, or slice each frame token's channels into per-part groups with separate projections and heads, relating parts through a fixed mixing matrix~\cite{zhang2023finemogen} or a per-frame attention branch over those slices~\cite{zhang2024lmm} rather than through part tokens. Concurrent work~\cite{zhang2026samor} finds eight learned part queries preferable to global and per-joint tokens for motion modelling over arbitrary skeletons. We place part tokens on the sequence axis of a single DiT, giving each hand, which carries $15$ of the $52$ joints, its own token and a nonlinear projector. Although coarser tokens are expected to degrade DiT quality~\cite{peebles2023dit}, that law concerns lossy patchification. A joint rotation is already a complete semantic unit, so anatomical grouping with part-specific decoders preserves per-joint resolution (Section~\ref{sec:exp_ablation}).

\section{Method}
\label{sec:method}

We formulate text-to-pose generation as Riemannian flow matching on the product manifold of SMPL-H joint rotations. Our approach combines two key components: (i) a geometry-aware flow-matching objective defined directly on $\SO^{52}$, which learns intrinsic tangent-space vector fields along geodesic paths, and (ii) a part-tokenized Diffusion Transformer that efficiently models whole-body articulation, including detailed hand articulation, through anatomical structure-aware tokenization.

\subsection{Pose Representation on \texorpdfstring{$\SO^{52}$}{SO(3)x52}}
\label{sec:method_manifold}

A SMPL-H pose is parameterised by 52 joint rotations: one global orientation, 21 body joint rotations, and $2 \times 15$ finger rotations. We represent each joint rotation as an element of the special orthogonal group $\SO$, and the full pose as an element of the product manifold $\mathcal{M} = \SO^{52}$. Articulated rotations naturally admit a Lie-group structure with well-defined geodesics and tangent spaces, making them suitable for intrinsic flow matching. For each joint, the Riemannian exponential and logarithm are
\begin{align}
    \Exp_{R}(v) &= R\, \exp([v]_\times), \\
    \Log_{R}(R') &= [\log(R^\top R')]_\vee,
\end{align}
where $[\cdot]_\times: \R^3 \to \so$ is the skew-symmetric mapping and $[\cdot]_\vee$ its inverse. The geodesic distance is $d_g(R, R') = \|\Log_{R}(R')\|_2$, and we extend these definitions to the product manifold by acting independently on each of the 52 joints. Importantly, the product-manifold formulation defines local rotational geometry but does not assume statistical independence between joints. Dependencies are modeled by the transformer vector field.

\subsection{Riemannian Flow Matching}
\label{sec:method_flow}

% Single-column (\columnwidth) manifold/geodesic figure for two-column WACV.
% SELF-CONTAINED: all helper macros (\px \py \amp, ff function) live INSIDE
% the tikzpicture, so nothing extra is needed in the preamble except:
%   \usepackage{tikz}\usepackage{amssymb}
%   \usetikzlibrary{arrows.meta,calc}
%   and your \Exp / \Log operator macros.
% (No decorations.markings / 3d libraries required.)
% Changes vs. the full-width version: legend moved from the right side to a
% compact one-row strip below the sheet; figure* -> figure; \textwidth ->
% \columnwidth. The picture is ~8.2cm wide, so at \columnwidth the fonts
% render at essentially their natural size.
\begin{figure}[t]
  \centering
  \resizebox{\columnwidth}{!}{%
  \begin{tikzpicture}[>=Latex,line join=round,line cap=round]

  % --- height field + oblique projection ---
  \def\px{0.46}   % y -> x shear
  \def\py{0.30}   % y -> screen-y shear (depth)
  \def\amp{0.55}  % surface amplitude

  \definecolor{sheet}{RGB}{150,170,205}
  \definecolor{sheetline}{RGB}{110,130,175}
  \definecolor{geod}{RGB}{200,40,40}        % geodesic = red
  \definecolor{eucl}{RGB}{225,150,40}       % euclidean = orange/amber
  \definecolor{tangent}{RGB}{40,140,90}     % velocity = green
  \definecolor{ptcol}{RGB}{30,30,40}

  \pgfmathdeclarefunction{ff}{2}{%
    \pgfmathparse{\amp*sin(90*#1)*cos(70*#2)}%
  }

  % ---------------------------------------------------------------------
  % 1) manifold sheet: wireframe + light fill
  % ---------------------------------------------------------------------
  \begin{scope}
  \foreach \y in {-2.4,-1.8,...,2.4}{
    \foreach \x in {-3.0,-2.4,...,2.4}{
      \pgfmathsetmacro{\xa}{\x}\pgfmathsetmacro{\xb}{\x+0.6}
      \pgfmathsetmacro{\ya}{\y}\pgfmathsetmacro{\yb}{\y+0.6}
      \pgfmathsetmacro{\za}{\amp*sin(90*\xa)*cos(70*\ya)}
      \pgfmathsetmacro{\zb}{\amp*sin(90*\xb)*cos(70*\ya)}
      \pgfmathsetmacro{\zc}{\amp*sin(90*\xb)*cos(70*\yb)}
      \pgfmathsetmacro{\zd}{\amp*sin(90*\xa)*cos(70*\yb)}
      \pgfmathsetmacro{\shade}{max(8,min(40,18+6*(\y)))}
      \fill[sheet!\shade!white,opacity=0.85]
        ({\xa+\px*\ya},{\za+\py*\ya}) --
        ({\xb+\px*\ya},{\zb+\py*\ya}) --
        ({\xb+\px*\yb},{\zc+\py*\yb}) --
        ({\xa+\px*\yb},{\zd+\py*\yb}) -- cycle;
    }
  }
  \foreach \yy in {-2.4,-1.8,...,2.4}{
    \draw[sheetline,very thin,opacity=0.55]
      plot[domain=-3.0:3.0,samples=40]
        ({\x+\px*\yy},{\amp*sin(90*\x)*cos(70*\yy)+\py*\yy});
  }
  \foreach \xx in {-3.0,-2.4,...,3.0}{
    \draw[sheetline,very thin,opacity=0.55]
      plot[domain=-2.4:2.4,samples=30,variable=\y]
        ({\xx+\px*\y},{\amp*sin(90*\xx)*cos(70*\y)+\py*\y});
  }
  \end{scope}

  % manifold label (moved up to tighten figure height)
  \node[sheetline!80!black,font=\small] at (3.1,-1.35) {$\mathcal{M}\subset\mathbb{R}^{n}$};
  \node[sheetline!80!black,font=\scriptsize] at (3.1,-1.70) {(valid rotations)};

  % helper macro: screen coords of a point (x,y) ON the sheet
  \newcommand{\sheetpt}[2]{%
    \pgfmathsetmacro{\sx}{#1+\px*#2}%
    \pgfmathsetmacro{\sy}{\amp*sin(90*#1)*cos(70*#2)+\py*#2}%
  }

  \def\xzero{-2.4}\def\yzero{-1.5}
  \def\xone{2.3}\def\yone{1.4}
  \def\xt{0.35}\def\yt{0.1}

  % ---------------------------------------------------------------------
  % 3) Euclidean chord (straight in ambient space), dashed, bowed upward
  % ---------------------------------------------------------------------
  \sheetpt{\xzero}{\yzero} \coordinate (P0) at (\sx,\sy);
  \sheetpt{\xone}{\yone}   \coordinate (P1) at (\sx,\sy);
  \coordinate (Cmid) at ($($(P0)!0.5!(P1)$)+(0,1.55)$);
  \draw[eucl,very thick,dash pattern=on 4pt off 2.5pt]
    (P0) .. controls (Cmid) and (Cmid) .. (P1);
  \coordinate (Eapex) at ($(P0)!0.5!(P1)!0.75!(Cmid)$);
  \draw[eucl,very thick,-{Latex[length=2.6mm]}]
    ($(Eapex)-(0.18,0)$) -- ($(Eapex)+(0.18,0)$);
  \draw[eucl!60,thin,dotted] (Cmid) -- ($(Cmid)+(0,-1.55)$);

  % ---------------------------------------------------------------------
  % 4) tangent plane patch + velocity at x_t
  % ---------------------------------------------------------------------
  \sheetpt{\xt}{\yt} \coordinate (Xt) at (\sx,\sy);
  \def\hh{0.18}
  \sheetpt{\xt+\hh}{\yt} \coordinate (Tx) at (\sx,\sy);
  \sheetpt{\xt}{\yt+\hh} \coordinate (Ty) at (\sx,\sy);
  \coordinate (e1) at ($(Tx)-(Xt)$);
  \coordinate (e2) at ($(Ty)-(Xt)$);
  \pgfmathsetmacro{\sc}{6.5}
  \fill[tangent!12,opacity=0.78]
    ($(Xt)+\sc*(e1)+\sc*(e2)$) --
    ($(Xt)+\sc*(e1)-\sc*(e2)$) --
    ($(Xt)-\sc*(e1)-\sc*(e2)$) --
    ($(Xt)-\sc*(e1)+\sc*(e2)$) -- cycle;
  \draw[tangent!55,thin]
    ($(Xt)+\sc*(e1)+\sc*(e2)$) --
    ($(Xt)+\sc*(e1)-\sc*(e2)$) --
    ($(Xt)-\sc*(e1)-\sc*(e2)$) --
    ($(Xt)-\sc*(e1)+\sc*(e2)$) -- cycle;
  \node[tangent!70!black,font=\scriptsize,anchor=south east]
    at ($(Xt)-\sc*(e1)+\sc*(e2)$) {$T_{x_t}\mathcal{M}$};

  % geodesic ON the sheet (red), drawn on top of the tangent patch
  \draw[geod,very thick]
    plot[domain=0:1,samples=60]
      ({(\xzero+(\xone-\xzero)*\x)+\px*(\yzero+(\yone-\yzero)*\x)},
       {\amp*sin(90*(\xzero+(\xone-\xzero)*\x))*cos(70*(\yzero+(\yone-\yzero)*\x))
         +\py*(\yzero+(\yone-\yzero)*\x)});
  \pgfmathsetmacro{\gmxa}{\xzero+(\xone-\xzero)*0.49}
  \pgfmathsetmacro{\gmya}{\yzero+(\yone-\yzero)*0.49}
  \pgfmathsetmacro{\gmxb}{\xzero+(\xone-\xzero)*0.55}
  \pgfmathsetmacro{\gmyb}{\yzero+(\yone-\yzero)*0.55}
  \sheetpt{\gmxa}{\gmya}\coordinate (Gma) at (\sx,\sy);
  \sheetpt{\gmxb}{\gmyb}\coordinate (Gmb) at (\sx,\sy);
  \draw[geod,very thick,-{Latex[length=2.6mm]}] (Gma) -- (Gmb);

  % velocity arrow: tangent to the geodesic at x_t, pointing toward x_1
  \def\dd{0.06}
  \pgfmathsetmacro{\gxa}{\xt-\dd*(\xone-\xzero)}\pgfmathsetmacro{\gya}{\yt-\dd*(\yone-\yzero)}
  \pgfmathsetmacro{\gxb}{\xt+\dd*(\xone-\xzero)}\pgfmathsetmacro{\gyb}{\yt+\dd*(\yone-\yzero)}
  \sheetpt{\gxa}{\gya}\coordinate (Ga) at (\sx,\sy);
  \sheetpt{\gxb}{\gyb}\coordinate (Gb) at (\sx,\sy);
  \coordinate (vel) at ($(Gb)-(Ga)$);
  \draw[tangent,very thick,->,>={Latex[length=2.6mm]}]
    (Xt) -- ($(Xt)+2.6*(vel)$)
    node[tangent!80!black,font=\small,anchor=south west,xshift=1pt,yshift=0pt]
    {$\dfrac{\Log_{x_t}(x_1)}{1-t}$};

  % ---------------------------------------------------------------------
  % 5) point markers + labels
  % ---------------------------------------------------------------------
  \foreach \P in {P0,P1,Xt}{
    \fill[ptcol] (\P) circle (2.2pt);
    \draw[white,line width=0.5pt] (\P) circle (2.2pt);
  }
  \node[ptcol,font=\small,anchor=north east] at ($(P0)+(-0.02,-0.05)$) {$x_0\sim p_0$};
  \node[ptcol,font=\small,anchor=north west] at ($(P1)+(0.05,-0.05)$)  {$x_1\sim p_1$};
  \node[ptcol,font=\small,anchor=north]      at ($(Xt)+(-0.15,-0.12)$) {$x_t$};

  % ---------------------------------------------------------------------
  % 6) legend: compact one-row strip BELOW the sheet (fits \columnwidth)
  % ---------------------------------------------------------------------
  \begin{scope}[shift={(0,-2.15)}]
    \draw[geod,very thick] (-4.05,0) -- (-3.55,0);
    \node[anchor=west,font=\footnotesize] at (-3.45,0) {geodesic on $\mathcal{M}$};
    \draw[eucl,very thick,dash pattern=on 4pt off 2.5pt] (-0.95,0) -- (-0.45,0);
    \node[anchor=west,font=\footnotesize] at (-0.35,0) {Euclidean chord};
    \draw[tangent,very thick,->,>={Latex[length=2mm]}] (1.95,0) -- (2.45,0);
    \node[anchor=west,font=\footnotesize] at (2.55,0) {tangent velocity};
  \end{scope}

  \end{tikzpicture}
  }
  \caption{Riemannian flow matching on the product manifold of joint rotations (schematic). Valid poses lie on a manifold $\mathcal{M}\subset\mathbb{R}^{n}$ embedded in the ambient representation space. Noise $x_0\sim p_0$ is transported to data $x_1\sim p_1$ along the on-manifold geodesic (Eq.~\ref{eq:interp}), whereas the Euclidean baseline follows the straight chord and leaves the manifold. At $x_t$ the network regresses the tangent velocity (Eq.~\ref{eq:target}), and integration stays on $\mathcal{M}$.}
  \label{fig:geodesic}
\end{figure}
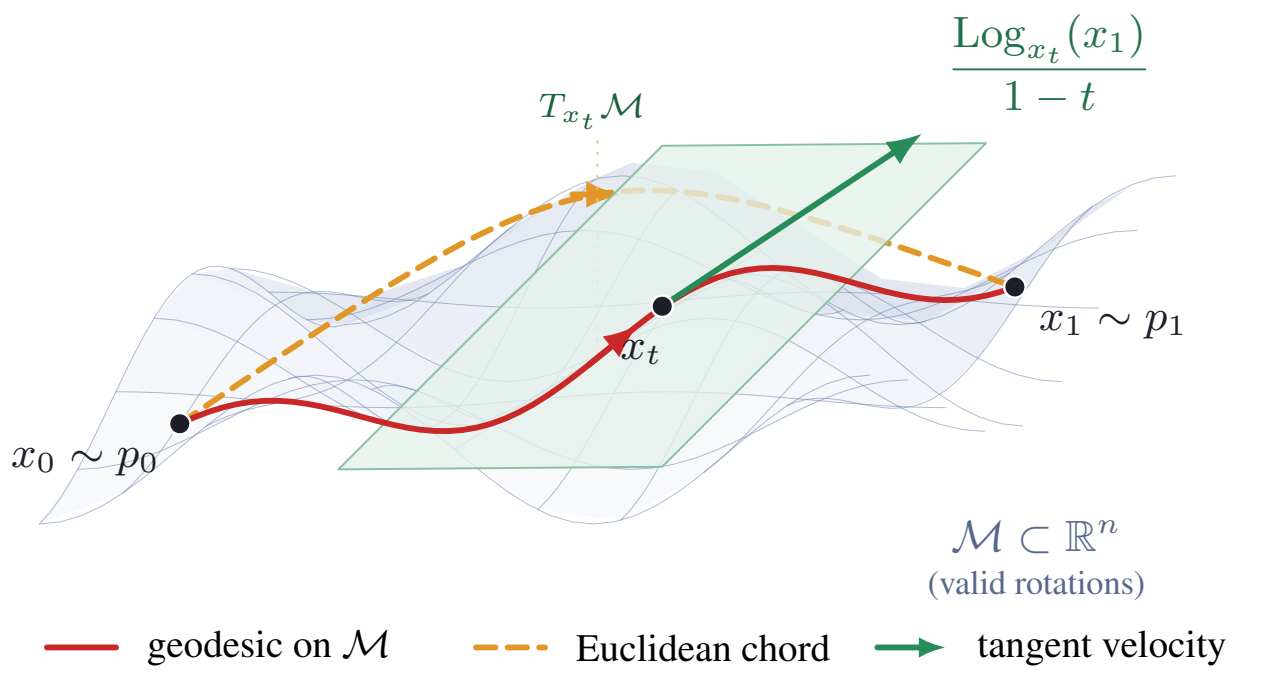
\begin{figure*}[t]
  \centering
  \resizebox{\textwidth}{!}{%
  \begin{tikzpicture}[
    >=Latex, line join=round, font=\sffamily,
    box/.style={rounded corners=2.5pt, line width=0.8pt, minimum height=8mm,
                inner sep=4pt, align=center, font=\small\sffamily},
    tok/.style={rounded corners=1.5pt, draw=mdC!55, fill=white, line width=0.5pt,
                minimum width=14mm, minimum height=3.6mm, font=\scriptsize\sffamily,
                text=mdC!25!black},
    otok/.style={rounded corners=1.5pt, draw=outC!55, fill=white, line width=0.5pt,
                 minimum width=14mm, minimum height=3.6mm, font=\scriptsize\sffamily,
                 text=outC!25!black},
    chip/.style={rounded corners=1.5pt, draw=mdC!55, fill=white, line width=0.6pt,
                 minimum width=30mm, minimum height=6mm, inner sep=1.5pt,
                 align=center, font=\footnotesize\sffamily, text=mdC!25!black},
    zchip/.style={rounded corners=2pt, inner sep=2.5pt, text=white,
                  font=\footnotesize\bfseries\sffamily},
    lbl/.style={font=\footnotesize\sffamily},
    tiny/.style={font=\fontsize{6}{7}\selectfont\sffamily},
    flow/.style={->, line width=1.0pt, draw=black!70},
    thin/.style={->, line width=0.6pt, draw=black!45},
    fan/.style={line width=0.5pt, draw=black!40},
    cflow/.style={->, line width=0.8pt, draw=black!50, dash pattern=on 3pt off 2pt},
  ]
  \definecolor{inC}{RGB}{50,100,160}
  \definecolor{mdC}{RGB}{115,85,165}
  \definecolor{outC}{RGB}{45,125,85}

  % ---- zones ----
  \draw[fill=inC!6,  draw=inC!30,  rounded corners=3pt, line width=0.5pt]
    (5mm,3mm)    rectangle (42.5mm,57.5mm);
  \draw[fill=mdC!6,  draw=mdC!30,  rounded corners=3pt, line width=0.5pt]
    (45.5mm,3mm) rectangle (166mm,57.5mm);
  \draw[fill=outC!6, draw=outC!30, rounded corners=3pt, line width=0.5pt]
    (169mm,26mm) rectangle (231.5mm,57.5mm);
  \node[zchip, fill=inC!85]  at (16mm,57.5mm)  {INPUTS};
  \node[zchip, fill=mdC!85]  at (58mm,57.5mm)  {MODEL};
  \node[zchip, fill=outC!85] at (182mm,57.5mm) {OUTPUT};

  % ---- inputs ----
  \node[box, draw=inC!75, fill=inC!12, text=inC!30!black, text width=26mm]
    (rt)  at (23.75mm,40mm)   {Noisy Pose $R_t$\\ \scriptsize 52 joints, 6D};
  \node[box, draw=inC!75, fill=inC!12, text=inC!30!black, text width=26mm]
    (txt) at (23.75mm,18.5mm) {Text Prompt};
  \node[box, draw=inC!75, fill=inC!12, text=inC!30!black, text width=26mm]
    (tt)  at (23.75mm,8.5mm)  {Timestep $t$};

  % ---- part projectors -> 8 part tokens ----
  \node[draw=mdC!75, fill=mdC!12, rounded corners=2.5pt, line width=0.8pt,
        minimum width=11mm, minimum height=22mm] (iproj) at (54.5mm,40mm) {};
  \node[rotate=90] at (iproj)
    {\scriptsize\sffamily\color{mdC!25!black} Part Projectors};
  \draw[flow] (rt.east) -- (iproj.west |- rt.east);
  \foreach \i/\nm/\yy in {0/torso/49,1/head/44.5,2/l-arm/40,3/r-hand/31.5}{
    \node[tok] (ti\i) at (70.5mm,\yy mm) {\nm};
    \draw[thin] (iproj.east |- ti\i) -- (ti\i.west);
  }
  \node[text=mdC!60] at (70.5mm,36.3mm) {$\vdots$};
  \node[lbl, text=black!55] at (70.5mm,53.5mm) {8 part tokens};
  \node[rounded corners=1.5pt, draw=mdC!70, dash pattern=on 2pt off 1.5pt,
        fill=white, line width=0.5pt, tiny, text=mdC!25!black,
        inner sep=2pt, minimum height=4.5mm]
    (ptemb) at (70.5mm,26mm) {$\oplus$\, part-type embedding};
  \draw[->, line width=0.6pt, draw=mdC!70, dash pattern=on 2pt off 1.5pt]
    (ptemb.north) -- (70.5mm,29.4mm);

  % ---- DiT block (components in a 2x2 grid) ----
  \node[draw=mdC!75, fill=mdC!10, rounded corners=3pt, line width=1pt,
        minimum width=68mm, minimum height=25mm] (blk) at (122mm,41.5mm) {};
  \node[font=\small\bfseries\sffamily, text=mdC!30!black] at (122mm,50.5mm)
    {DiT Block $\times N$};
  \node[chip] at (105.5mm,43.5mm) {AdaLN$(t)$};
  \node[chip] at (138.5mm,43.5mm) {Self-Attention};
  \node[chip] at (105.5mm,35.5mm)
    {Cross-Attention\\[-1pt] \fontsize{6}{7}\selectfont (text $K,V$)};
  \node[chip] at (138.5mm,35.5mm) {SwiGLU FFN};

  % ---- token bus in / out ----
  \coordinate (busIn) at (84mm,40mm);
  \foreach \i in {0,...,3}{ \draw[fan] (ti\i.east) -- (busIn); }
  \fill[black!60] (busIn) circle (0.7mm);
  \draw[flow] (busIn) -- (blk.west |- busIn);
  \coordinate (busOut) at (160mm,40mm);
  \draw[flow] (blk.east |- busOut) -- (busOut);
  \fill[black!60] (busOut) circle (0.7mm);

  % ---- conditioning (encoders inside MODEL, dashed into block bottom) ----
  \node[box, draw=mdC!45, fill=white, text=black!60, text width=22mm,
        font=\footnotesize\sffamily, minimum height=9mm]
    (tenc) at (63.25mm,18.5mm) {Flan-T5 Encoder\\ \scriptsize frozen, all $L$ tokens};
  \node[box, draw=mdC!45, fill=white, text=black!60, text width=27mm,
        font=\scriptsize\sffamily, minimum height=8mm]
    (temb) at (63.25mm,8.5mm) {Time Emb.\\ sinusoidal$(t)$+MLP};
  \draw[thin] (txt.east) -- (tenc.west);
  \draw[thin] (tt.east)  -- (temb.west);
  \draw[cflow] (tenc.east) -| (105.5mm,29mm);
  \draw[cflow] (temb.east) -| (138.5mm,29mm);
  \node[lbl, text=black!50, anchor=south] at (88mm,19mm) {$K,V$};
  \node[lbl, text=black!50, anchor=south] at (128mm,9mm) {$t$};

  % ---- output ----
  \foreach \i/\nm/\yy in {0/torso/49,1/head/44.5,2/l-arm/40,3/r-hand/31.5}{
    \node[otok] (to\i) at (181.5mm,\yy mm) {\nm};
    \draw[thin] (busOut) -- (to\i.west);
  }
  \node[text=outC!60] at (181.5mm,36.3mm) {$\vdots$};
  \node[lbl, text=black!55] at (181.5mm,53.5mm) {denoised tokens};
  \node[draw=outC!75, fill=outC!12, rounded corners=2.5pt, line width=0.8pt,
        minimum width=11mm, minimum height=22mm] (odec) at (197mm,40mm) {};
  \node[rotate=90] at (odec)
    {\scriptsize\sffamily\color{outC!25!black} Part Decoders};
  \foreach \i in {0,...,3}{ \draw[thin] (to\i.east) -- (odec.west |- to\i); }
  \node[box, draw=outC!75, fill=outC!12, text=outC!25!black, text width=18mm]
    (vel) at (218.5mm,40mm) {Velocity $\hat v_\theta$\\ \scriptsize 3D tangent/joint};
  \draw[flow] (odec.east) -- (vel.west);

  \end{tikzpicture}
  }
    \caption{\textbf{PoseRieFlow architecture.} Part-specific input projectors (linear for body parts, MLPs for the hands) map the noisy pose $R_t$ (52 joints, 6D-encoded) to eight anatomical part tokens with part-type embeddings. Each DiT block applies self-attention over the part tokens, cross-attention to frozen Flan-T5-Base text features, and a SwiGLU FFN, modulated by AdaLN conditioned on the timestep with zero-initialised gates. Mirrored decoders map each token back to its joints, yielding the per-joint tangent velocity $\hat{v}_\theta$ (Sec.~\ref{sec:method_arch}).}
  \label{fig:architecture}
\end{figure*}
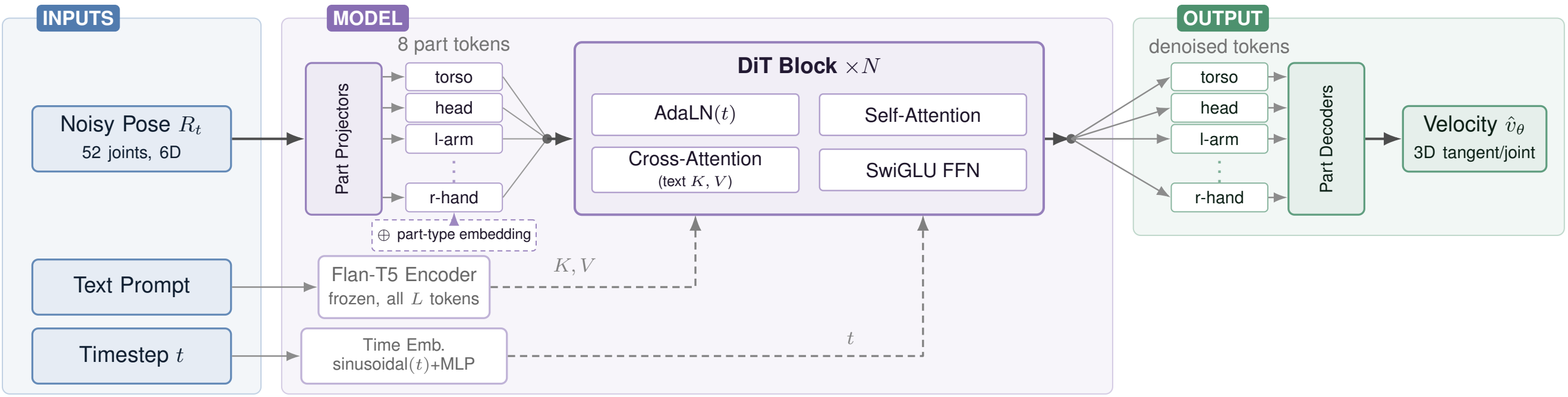

Flow matching trains a time-dependent vector field $v_\theta$ to transport samples from a tractable prior $p_0$ to the data distribution $p_1$ along a prescribed interpolation path. Unlike Euclidean flow matching, which learns velocities in an ambient parameter space, we learn intrinsic tangent-space velocities on the manifold of rotations. On $\SO$, the prior $p_0$ on each joint is an isotropic Gaussian in the Lie algebra mapped to the group by the exponential map: we draw an axis-angle vector $\omega \sim \mathcal{N}(0, \sigma^2 I_3)$ in $\so \cong \R^3$ and set $R_0 = \Exp_{I}(\omega) = \exp([\omega]_\times)$, with $\sigma = 1$ radian throughout. The full-pose prior is the product of 52 such independent draws. Given a data sample $R_1$ and a prior sample $R_0 \sim p_0$, we connect them through the geodesic interpolant
\begin{equation}
    R_t = \Exp_{R_0}\!\big(t \cdot \Log_{R_0}(R_1)\big), \quad t \in [0,1].
    \label{eq:interp}
\end{equation}
Unlike linear interpolation in Euclidean rotation parameterizations, the geodesic interpolant remains on the manifold and follows the shortest rotational path between poses.

\paragraph{Velocity Target.}
Rather than regressing Euclidean rotation updates, the model predicts the transport direction in the local tangent space of each joint. Writing the remaining rotation in the body frame at $R_t$ as $\omega = \Log_{R_t}(R_1) = [\log(R_t^\top R_1)]_\vee \in \R^3$, the geodesic velocity is
\begin{equation}
    v_\text{target}(R_t, R_1, t) = \frac{\omega}{1 - t} = \frac{\Log_{R_t}(R_1)}{1 - t},
    \label{eq:target}
\end{equation}
where $\omega$ is the remaining rotation and $(1-t)$ is the remaining time to reach $R_1$. The induced body-frame dynamics are $\tfrac{d}{dt}R_t = R_t\,[\omega / (1-t)]_\times$. The network predicts a 3-dimensional tangent vector per joint, $\hat{v}_{\theta,j}(x_t, t, c) \in \R^3$, for $j = 1, \ldots, 52$, matching this target directly.  Similar to other pose and motion generation models we encode the current point $R_t$ using the continuous 6D rotation representation \cite{zhou20196d}, which we denote $ x_t $, as input for the model, while all supervision and integration operate in tangent space.

\paragraph{Loss.}
We supervise the predicted tangent vectors with a mean-squared error objective on the product manifold,
\begin{equation}
    \label{eq:loss}
    \mathcal{L}(\theta) = \mathbb{E}_{t, R_0, R_1, c}\!\left[
    %\frac{1}{52} 
    \sum_{j=1}^{52} \big\| \hat{v}_{\theta,j}(x_t, t, c) - v_{\text{target},j} \big\|_2^2 \right].
\end{equation}
The expectation is over $t \sim \mathcal{U}[0, 0.999]$, $R_0$ from the Riemannian-Gaussian prior, and $R_1$ with paired caption $c$ from the data. Operating in the tangent space yields gradient directions aligned with the geodesic structure of articulated rotations, in contrast to Euclidean flow matching that interpolates and regresses in $\R^6$. Because geodesic interpolation follows valid rotational trajectories, the resulting vector field is more consistent with the pose geometry. For comparison, our Euclidean baseline replaces the manifold interpolant with linear interpolation in 6D space,
\[
x_t = (1-t)x_0 + tx_1,
\]
where $x_0 \sim \mathcal{N}(0,I)$, and regresses the constant velocity $x_1 - x_0$. In this setting the network predicts $\hat{v}_\theta \in \R^6$ and is trained with a standard 6D MSE objective.

\subsection{Network Architecture}
\label{sec:method_arch}

\paragraph{Part Tokenization.}
Naively treating each SMPL-H joint as an independent transformer token increases sequence length and attention cost, while weakening the modeling of correlated local structure, particularly in the hands. Instead, following part-token designs in discriminative skeleton and pose models~\cite{wang2023iip,dou2023tore,wang2024gtpt}, we aggregate joints into eight semantic body-part tokens: torso (5 joints), head (1), left arm (4), right arm (4), left leg (4), right leg (4), left hand (15), and right hand (15). This granularity provides a favorable trade-off between attention efficiency and preservation of local articulation structure, particularly for dexterous hand articulation. Each part is mapped to a single $d$-dimensional token by a part-specific projector applied to its concatenated joint inputs. Body parts use a single linear layer, while the two hands use a one-hidden-layer MLP with GELU activation to better model finger coordination and higher articulation complexity. A learned part-identity embedding is added to each part token. The eight part tokens form the transformer input sequence. After transformer processing, a part-specific decoder maps each part token back to its constituent joints (linear for body parts and a one-hidden-layer MLP for hands). The resulting per-joint outputs are scattered back into the canonical SMPL-H joint ordering. The output dimension is 3 per joint in the Riemannian setting (tangent vectors) and 6 in the Euclidean ablation.

\paragraph{Text Conditioning.}
We condition on natural language through cross-attention to a frozen Flan-T5-Base~\cite{chung2024t5} encoder. For a given caption we extract the per-token features
\[
\mathbf{E}_\text{text}
=
[\,\mathbf{e}_1, \dots, \mathbf{e}_L\,]
\in \R^{L \times 768}
\]
and project them once to the model dimension, $\tilde{\mathbf{E}} = \mathrm{LN}(\mathbf{E}_\text{text}\,W_\text{text}) \in \R^{L \times d}$. At 250M parameters the encoder is an order of magnitude smaller than the billion-scale text and vision-language backbones used by prior text-to-pose systems, yet sufficient for strong text-pose alignment. A pooled-conditioning variant and a larger $4$B embedding encoder are ablated in Table~\ref{tab:cond} and shown to be inferior.

\paragraph{Diffusion Transformer Blocks.}
The part-token sequence $\mathbf{S}$ is processed by a stack of transformer blocks with adaptive LayerNorm (AdaLN) time conditioning. Each block comprises three sublayers. (i) \emph{Self-attention} over the eight part tokens, which models long-range dependencies such as coordinated hand, arm, and torso configurations. (ii) \emph{Cross-attention}, in which the part tokens attend to the text features $\tilde{\mathbf{E}}$, giving each part direct token-level access to the caption. (iii) A \emph{SwiGLU} feed-forward network. The scale, shift, and gate parameters for all three sublayers are conditioned on the sinusoidal timestep embedding. The gates are zero-initialised so that each block initially behaves as the identity, which stabilizes optimization during training.

\subsection{Implementation Details}
\label{sec:method_impl}

During training, we apply classifier-free guidance dropout with probability $0.1$, replacing the text embedding with a learned null embedding optimized jointly with the model. At inference, classifier-free guidance is applied as
\[
v_\theta^{\text{cfg}}
=
v_\theta(\emptyset)
+
w \cdot
\big(
v_\theta(\mathbf{e}_\text{text})
-
v_\theta(\emptyset)
\big),
\]
where $w$ denotes the guidance scale. We train using AdamW with learning rate $5 \times 10^{-4}$ and a warmup-cosine learning-rate schedule. Sampling integrates the predicted vector field along geodesics on $\SO$ using an explicit Euler scheme. Given the current state $R_t$, the network predicts a body-frame tangent vector $\hat{v}_\theta$, and updates
\[
R_{t+\Delta t}
\leftarrow
R_t \exp([\hat{v}_\theta \Delta t]_\times),
\]
with uniform step size $\Delta t = 1/N$. Sampling begins from the $\SO$ Riemannian-Gaussian prior and proceeds for $N=50$ steps with classifier-free guidance weight $w=2.0$. For fair comparison, all reported Euclidean and Riemannian results use identical NFE budgets unless stated otherwise. Further optimization settings and numerical-stability measures (mixed-precision handling and clamping in the logarithm and Gram-Schmidt maps) are provided in the supplementary material.

\section{Experiments}
\label{sec:experiments}

\begin{figure*}[t]
    \centering
    \setlength{\tabcolsep}{1pt}
    \renewcommand{\arraystretch}{0.4}
    \newcommand{\posefig}[1]{%
    \includegraphics[width=0.135\textwidth, trim=230 70 230 110, clip]{#1}}
    \begin{tabular}{@{}ccccccc@{}}
        \posefig{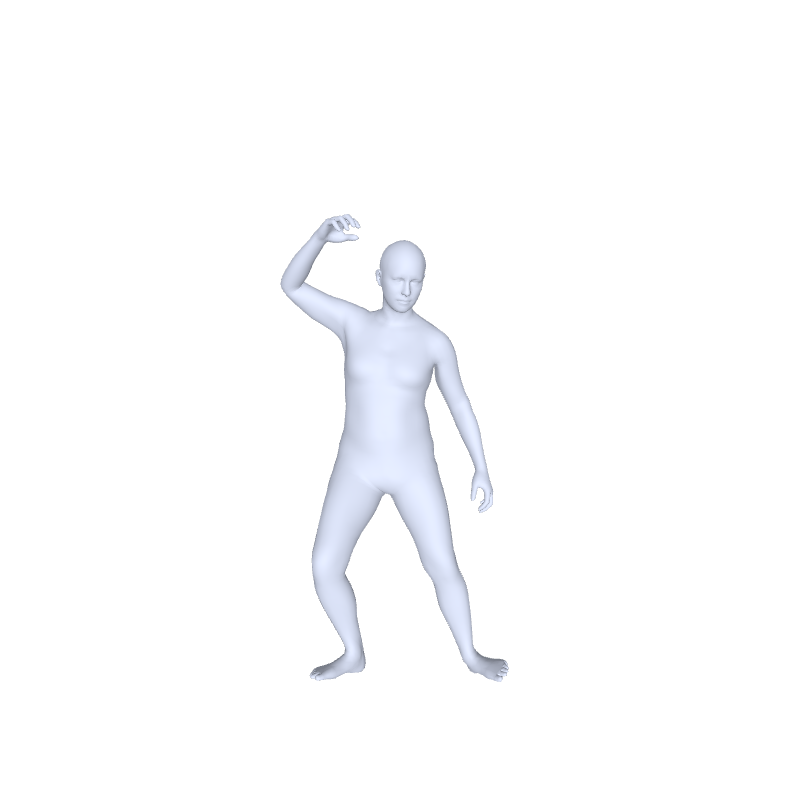} &
        \posefig{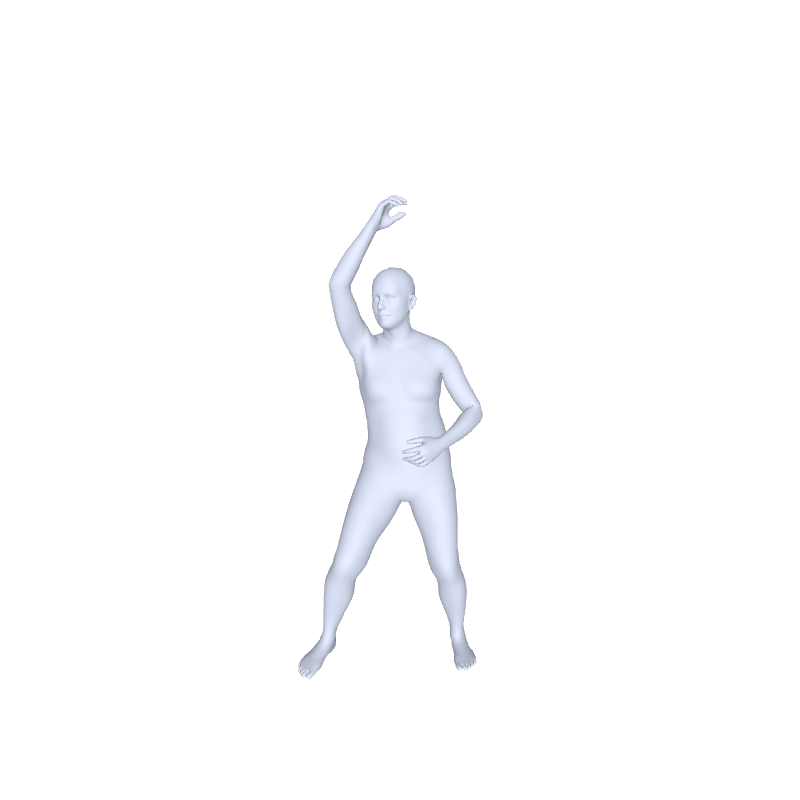} &
        \posefig{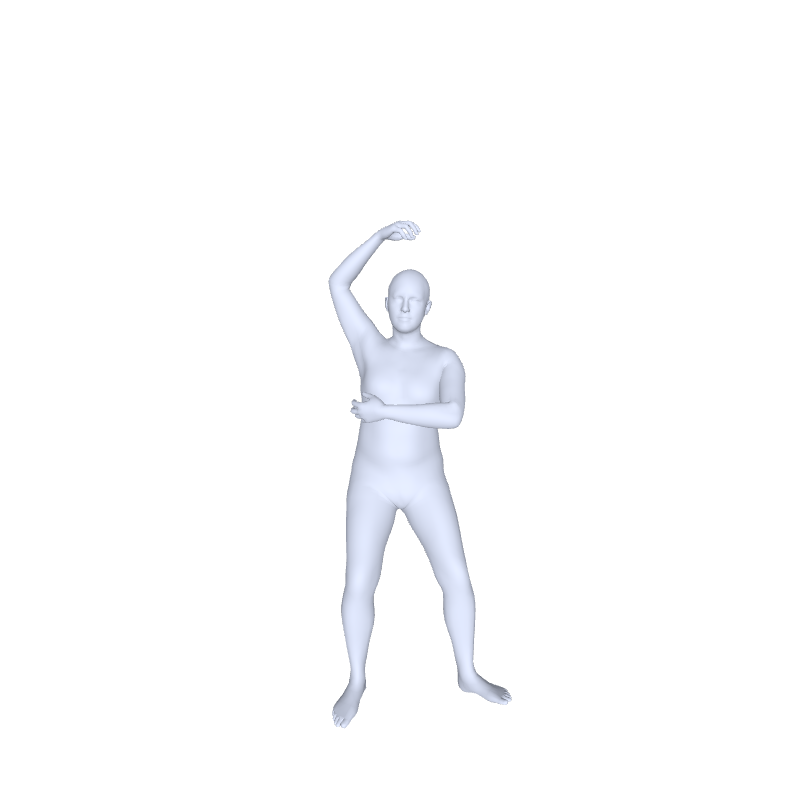} &
        \posefig{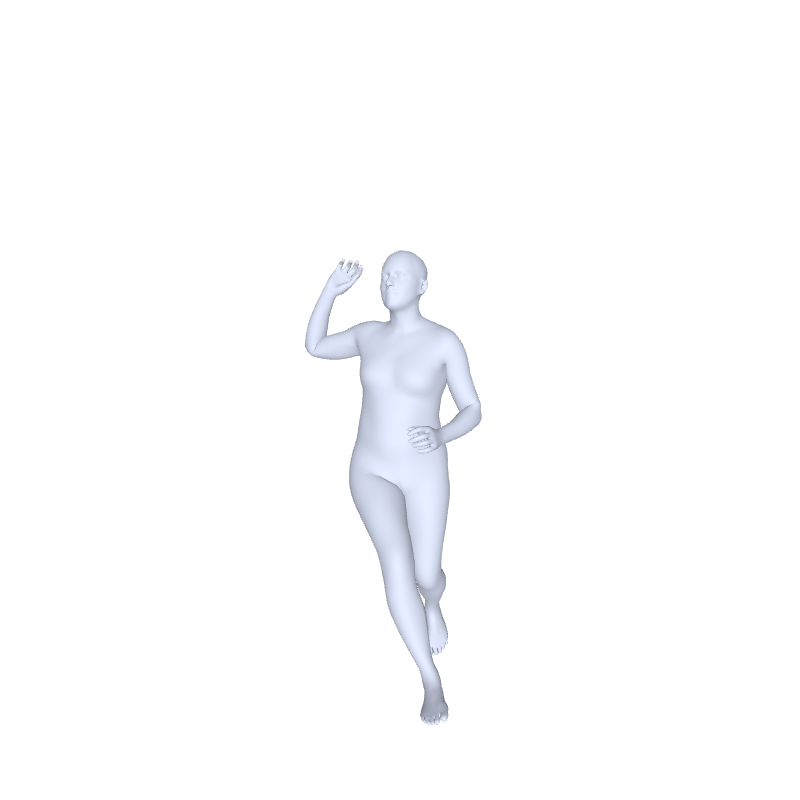} &
        \posefig{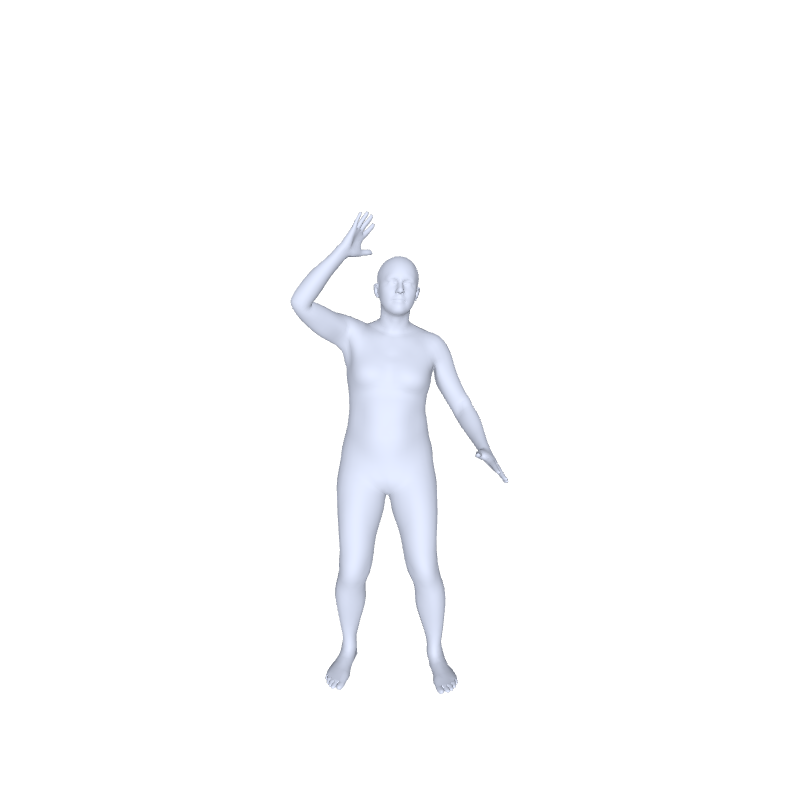} &
        \posefig{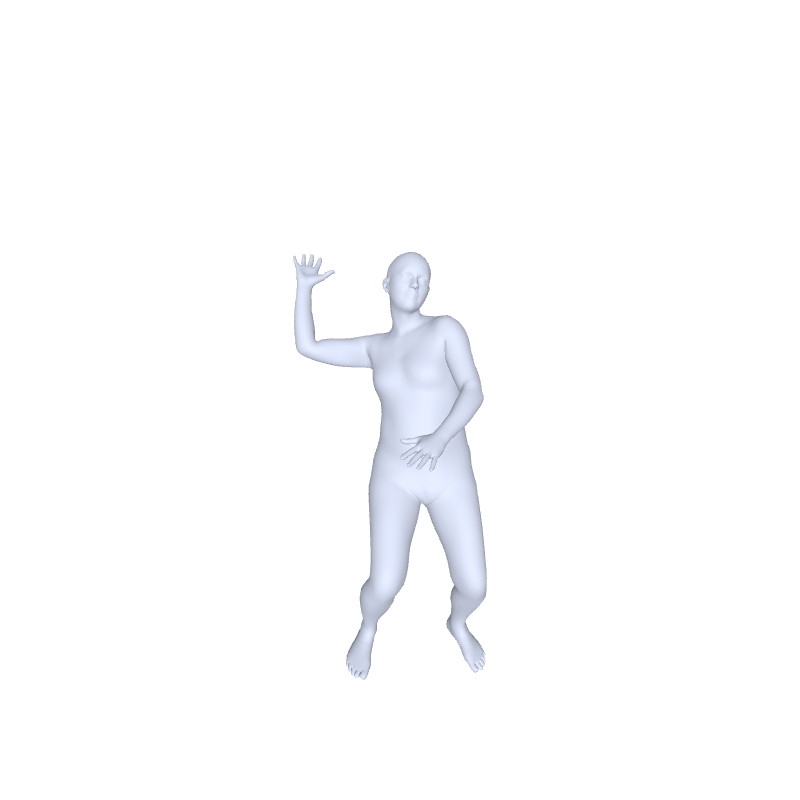} &
        \posefig{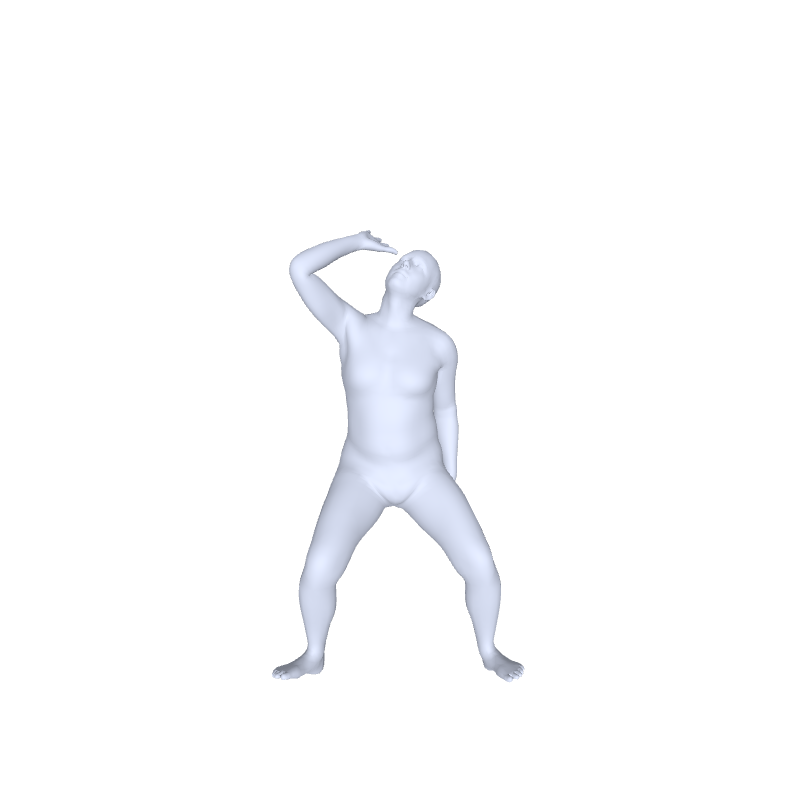} \\[-1pt]
        {\scriptsize Reference} &
        {\scriptsize \textbf{Ours (Riem.)}} &
        {\scriptsize Ours (Eucl.)} &
        {\scriptsize PoseScript} &
        {\scriptsize ChatPose} &
        {\scriptsize PoseLLaVA} &
        {\scriptsize UniPose} \\
    \end{tabular}
    \caption{\textbf{Qualitative whole-body generation.} \emph{``The right hand is reaching up, the right elbow is in l-shape and the left arm is beneath the right arm and the left elbow is bent a bit. The right knee is located in front of the left knee, rather bent and the torso is upright with the left hand located behind the right hand.''} PoseRieFlow and the PoseScript VAE produce articulated SMPL-H hands. The body-only LLM baselines emit flat, fingerless hands. Headings aligned as in Sec.~\ref{sec:exp_setup}.}
    \label{fig:qualitative}
\end{figure*}

We evaluate PoseRieFlow along two complementary axes. We first report results on the standard body-only benchmark used by prior text-to-pose work, where all baselines are applicable (Section~\ref{sec:exp_body}). We then evaluate full SMPL-H whole-body generation including articulated hands (Section~\ref{sec:exp_wholebody}), the primary target setting of PoseRieFlow. In both regimes we isolate the effect of the proposed manifold formulation through matched-compute comparisons between Riemannian and Euclidean flow matching, and we further analyze inference efficiency, guidance, tokenization, conditioning, and training data in Section~\ref{sec:exp_ablation}.

\subsection{Experimental Setup}
\label{sec:exp_setup}

\paragraph{Data.}
We train and evaluate on PoseScript-A2H2, the union of the PoseScript automatic (A2) and human-written (H2) caption splits, which pairs SMPL-H poses from AMASS with natural-language descriptions. Our default models are trained on PoseScript-A2H2 alone, matching the data available to the PoseScript VAE. A training-data ablation (Section~\ref{sec:exp_ablation}) additionally trains on a 300k-pose subset of Motion-X++~\cite{zhang2025motion}, a whole-body motion dataset whose frame-level captions provide detailed hand articulation that PoseScript does not emphasise. Baseline results use the authors' publicly released pretrained models. All methods are evaluated identically on the test split ($N = 19990$).

\paragraph{Metrics.}
All generated and reference poses are first rotated about the vertical axis to a common heading. We measure text-pose alignment with the PoseScript retrieval model \cite{delmas2024posescript} and report R@1 in a 32-candidate gallery, the correct item plus 31 seeded distractors, with identical galleries for all methods, a chance level of $3.125\,\%$, and exact ties counted as misses, in both the text-to-pose (T2P) and pose-to-text (P2T) directions, together with Fr\'echet Inception Distance (FID) in the retrieval model's pose-embedding space as a measure of distributional realism. We report FID in the whole-body setting (Section~\ref{sec:exp_wholebody}), where it is computed on complete 52-joint poses. In the body-only projection, zeroing the hand joints degenerates the retrieval feature covariance and renders the FID estimate unstable, so we omit it there. We additionally report per-joint kinematic error against the reference pose: Procrustes-aligned mean per-joint position error (PA-MPJPE) in millimetres and mean geodesic rotation error (Geo) in degrees, where $\mathrm{Geo}_{\setminus 0}$ denotes the same error with the global orientation excluded. In the whole-body setting we further report Geo restricted to the 30 finger joints and fingertip PA-MPJPE summed over the ten fingertips (TipPA). Stochastic methods are sampled with ten seeds per caption and reported as mean$\pm$std, and their diversity (Div) is the mean pairwise PA-MPJPE between multiple samples, averaged over the test set.

\paragraph{Baselines.}
We compare against the PoseScript conditional VAE \cite{delmas2024posescript}, the recent LLM-based generators ChatPose~\cite{feng2024chatpose}, PoseLLaVA~\cite{posellava2025}, and UniPose~\cite{li2025unipose}. PoseRieFlow is reported in two variants that share architecture, parameter count, and training budget and differ only in the flow formulation: \emph{Ours (Riem.)}, the Riemannian model with geodesic interpolation and tangent-space targets, and \emph{Ours (Eucl.)}, the Euclidean ablation that interpolates and regresses in $6$D. This pairing isolates the contribution of the manifold geometry at matched compute. Recent text-to-pose systems increasingly build upon large vision--language models containing several billion parameters ($13.43$B for ChatPose, $8.23$B for UniPose, and $7.64$B for PoseLLaVA), whereas PoseRieFlow uses a standalone 21.6M-parameter generator. We therefore compare not only generation quality but also the effectiveness of geometry-aware whole-body modeling at substantially smaller scale.

\subsection{Body-Only Pose Generation}
\label{sec:exp_body}

\begin{table*}[t]
    \centering
    \caption{Text-to-pose generation on PoseScript-A2H2 (body-only: 22 SMPL-H joints, predicted hands zeroed). Stochastic methods (ours at NFE$=$50, $w{=}2$, final checkpoint, PoseScript from its VAE prior) report mean$_{\pm\mathrm{std}}$ over 10 seeds, while the MLLM baselines decode greedily (Div$\,{=}\,0$). Shading: best (red), second (orange), and third (yellow) per column. P2T for every method sits at the chance floor.}
    \footnotesize
    \label{tab:main_smpl}
    \begin{tabular}{lcccccc}
        \toprule
         & \multicolumn{2}{c}{R@1(32) $\uparrow$ (\%)} & PA-MPJPE $\downarrow$ & Geo $\downarrow$ & Geo$_{\setminus 0}$ $\downarrow$ & Div \\
        \cmidrule(lr){2-3}
        Method & T2P & P2T & (mm) & ($^\circ$) & ($^\circ$) & (mm) \\
        \midrule
        PoseScript~(VAE) & $3.29_{\pm 0.09}$ & $3.17_{\pm 0.07}$ & $151.44_{\pm 0.14}$ & $30.06_{\pm 0.02}$ & $27.00_{\pm 0.02}$ & $462.0$ \\
        ChatPose & \cthird{$13.56$} & $3.27$ & $108.62$ & \cbest{$19.27$} & \cbest{$19.59$} & $0$ \\
        UniPose & \cbest{$22.61$} & $3.27$ & \cthird{$108.46$} & \csecond{$22.94$} & $23.49$ & $0$ \\
        PoseLLaVA & \csecond{$14.38$} & $3.19$ & $123.02$ & $26.96$ & $24.01$ & $0$ \\
        \midrule
        Ours (Eucl.) & $13.20_{\pm 0.17}$ & $3.25_{\pm 0.02}$ & \csecond{$80.72_{\pm 0.08}$} & $24.82_{\pm 0.01}$ & \cthird{$20.73_{\pm 0.01}$} & $181.9$ \\
        Ours (Riem.) & $13.15_{\pm 0.21}$ & $3.23_{\pm 0.01}$ & \cbest{$76.09_{\pm 0.08}$} & \cthird{$24.04_{\pm 0.02}$} & \csecond{$19.93_{\pm 0.02}$} & $222.2$ \\
        \bottomrule
    \end{tabular}
\end{table*}

We first evaluate in the body-only setting, where every baseline is applicable. Methods that produce SMPL-H output (PoseScript and PoseRieFlow) have their hand joints zeroed for this comparison, and all methods, regardless of how they internally parameterise pose, are scored on the same body skeleton with the same PoseScript retrieval model, so the metrics are directly comparable. Table~\ref{tab:main_smpl} reports the results. On Procrustes-aligned joint error, PoseRieFlow recovers body configurations closer to the reference than every baseline ($76.1$\,mm for the Riemannian variant versus $108.5$\,mm for the best LLM-based generator and $151.4$\,mm for the PoseScript VAE), despite PoseRieFlow being trained on the full SMPL-H manifold and projected down to the body joints for this comparison. On geodesic error, both flow variants improve upon PoseScript and PoseLLaVA, while ChatPose and UniPose attain lower values in this projection. Pose-to-text R@1 sits at the $3.125\,\%$ chance floor for every method, while text-to-pose retrieval favours the billion-parameter MLLMs ($22.61\,\%$ for UniPose versus $13.15\,\%$ for our Riemannian variant), consistent with their language-space pretraining. PoseRieFlow models the complete SMPL-H articulation, including both hands, and is subsequently projected onto the body skeleton for comparison. We therefore regard the whole-body evaluation in Section~\ref{sec:exp_wholebody} as the more representative assessment of the proposed formulation.

\subsection{Whole-Body Pose Generation with Hands}
\label{sec:exp_wholebody}

\begin{table*}[t]
    \centering
    \caption{Whole-body text-to-pose generation on PoseScript-A2H2 (all 52 SMPL-H joints). The body-only MLLM baselines are excluded. Every cell is mean$_{\pm\mathrm{std}}$ over 10 seeds, sampling settings as in Table~\ref{tab:main_smpl}. Shading: best (red), second (orange), and third (yellow) per column. In Hand Geo the two flow variants are tied. P2T is informative here, only PoseScript sits at the chance floor.}
    \footnotesize
    \label{tab:smplh}
    \begin{tabular}{lcccccccc}
        \toprule
         & & \multicolumn{2}{c}{R@1(32) $\uparrow$ (\%)} & PA-MPJPE $\downarrow$ & Geo $\downarrow$ & \multicolumn{2}{c}{Hands} & Div \\
        \cmidrule(lr){3-4}\cmidrule(lr){7-8}
        Method & FID $\downarrow$ & T2P & P2T & (mm) & ($^\circ$) & Geo $\downarrow$ ($^\circ$) & TipPA $\downarrow$ (mm) & (mm) \\
        \midrule
        PoseScript~(VAE) & \cthird{$1.691_{\pm 0.001}$} & \cthird{$3.29_{\pm 0.09}$} & \cthird{$3.17_{\pm 0.07}$} & \cthird{$206.54_{\pm 0.31}$} & \cthird{$26.23_{\pm 0.02}$} & \cthird{$23.410_{\pm 0.030}$} & \cthird{$207.15_{\pm 0.36}$} & $588.5$ \\
        \midrule
        Ours (Eucl.) & \cbest{$0.904_{\pm 0.002}$} & \cbest{$27.79_{\pm 0.16}$} & \cbest{$22.87_{\pm 0.21}$} & \csecond{$116.37_{\pm 0.19}$} & \csecond{$10.65_{\pm 0.01}$} & \cbest{$0.260_{\pm 0.003}$} & \csecond{$13.85_{\pm 0.02}$} & $254.1$ \\
        Ours (Riem.) & \csecond{$0.944_{\pm 0.002}$} & \csecond{$27.18_{\pm 0.12}$} & \csecond{$22.11_{\pm 0.14}$} & \cbest{$111.52_{\pm 0.14}$} & \cbest{$10.32_{\pm 0.01}$} & \cbest{$0.265_{\pm 0.001}$} & \cbest{$13.38_{\pm 0.02}$} & $297.6$ \\
        \bottomrule
    \end{tabular}
\end{table*}

A central objective of PoseRieFlow is the generation of full SMPL-H articulation, including the 30 finger joints omitted by most existing text-to-pose methods. To evaluate this setting, all retrieval-based metrics are recomputed for all 52 joints. To our knowledge, PoseScript is the only prior text-to-pose method that produces SMPL-H output and provides a directly comparable baseline in this setting, as existing LLM-based approaches operate exclusively on body joints. Table~\ref{tab:smplh} therefore evaluates the methods applicable to whole-body generation. Across the full 52-joint skeleton, PoseRieFlow attains the lowest PA-MPJPE among all applicable methods, with the Riemannian variant reaching $111.5$\,mm compared to $116.4$\,mm for the Euclidean variant and $206.5$\,mm for the PoseScript VAE. Against its matched Euclidean counterpart, the Riemannian variant improves both kinematic metrics (PA-MPJPE $111.5$ versus $116.4$\,mm, Geo $10.32^\circ$ versus $10.65^\circ$), and the advantage remains consistent across all practical inference budgets (Section~\ref{sec:exp_ablation}), suggesting that the gain is systematic rather than an artifact of a particular discretization. On the hands, both flow variants improve over the VAE by an order of magnitude or more (hand Geo $0.26^\circ$ versus $23.4^\circ$, TipPA $13.4$ to $13.9$ versus $207$\,mm), while between the variants fingertip PA-MPJPE improves from $13.85$ to $13.38$\,mm and hand geodesic error is statistically tied (Table~\ref{tab:smplh}). A2H2 ground-truth hands are nearly identical across items, so these metrics chiefly reward producing plausible hands, which the VAE frequently fails to do. Distributional and retrieval metrics do not follow the kinematic trend: the Euclidean variant is marginally better on FID ($0.904$ versus $0.944$) and in both retrieval directions ($27.8$ versus $27.2\,\%$ T2P), while both flow variants improve upon the PoseScript VAE on every metric by a wide margin, with whole-body T2P retrieval roughly eight times higher ($27.2\,\%$ versus $3.3\,\%$). Geometry-aware transport therefore buys kinematic fidelity at matched compute rather than uniformly improving every axis. 

\subsection{Ablations}
\label{sec:exp_ablation}

\begin{table}[t]
  \centering
  \caption{\textbf{Kinematic fidelity vs.\ sampling budget (NFE)} (all 52 joints, seed 0). The Riemannian variant is better on both metrics at every practical budget ($\mathrm{NFE}\geq5$), and the ordering reverses at $\mathrm{NFE}\leq2$ (Sec.~\ref{sec:exp_ablation}). Bold: better variant per budget. Full sweep incl.\ FID, retrieval, and hand metrics in the supplement.}
  \label{tab:nfe_compact}
  \footnotesize
  \setlength{\tabcolsep}{2pt}
  \begin{tabular}{@{}llccccccc@{}}
    \toprule
    & NFE & 1 & 2 & 5 & 10 & 20 & 50 & 100 \\
    \midrule
    \multirow{2}{*}{PA (mm)\,$\downarrow$}
      & Eucl. & \textbf{167.7} & \textbf{143.7} & 113.4 & 112.9 & 114.7 & 116.4 & 117.1 \\
      & Riem. & 213.6 & 156.4 & \textbf{111.5} & \textbf{109.3} & \textbf{110.4} & \textbf{111.6} & \textbf{112.1} \\
    \midrule
    \multirow{2}{*}{Geo (deg)\,$\downarrow$}
      & Eucl. & \textbf{14.85} & \textbf{12.38} & 10.69 & 10.57 & 10.61 & 10.65 & 10.69 \\
      & Riem. & 18.72 & 13.04 & \textbf{10.52} & \textbf{10.30} & \textbf{10.28} & \textbf{10.31} & \textbf{10.33} \\
    \bottomrule
  \end{tabular}
\end{table}

\subsubsection{Riemannian vs.\ Euclidean Flow}

Our Riemannian and Euclidean variants share architecture, parameter count, training schedule, and inference budget, differing only in the interpolation path and regression target (Section~\ref{sec:method_flow}). This makes the comparison the most direct test of the proposed contribution, isolating the effect of respecting rotational geometry from other factors. At the default budget of $N=50$ steps the Riemannian variant already attains lower geodesic rotation error than its Euclidean counterpart in both the body-only (Table~\ref{tab:main_smpl}) and whole-body (Table~\ref{tab:smplh}) settings, confirming that aligning the training signal with the geodesic structure of $\mathrm{SO}(3)$ improves rotational fidelity rather than changing the parameterisation. We further examine how this advantage depends on the inference budget by sweeping the number of function evaluations $N$ for both variants (Table~\ref{tab:nfe_compact}). Across the practical range \(N\in[5,100]\), the Riemannian variant achieves lower geodesic error by \(0.17^{\circ}\) to \(0.36^{\circ}\). In the one- and two-step regime (\(N\leq2\)), performance degrades sharply for both variants, and the ordering reverses: the geodesic prior's wide angular support makes a single Euler step on \(\mathrm{SO}(3)^{52}\) a poor geodesic approximation. Thus, PoseRieFlow yields rotational gains at moderate-to-low NFE (\(N\geq5\)) rather than in few-step settings. These gains are far outside sampling noise. Over ten seeds, per-seed standard deviations (Tables~\ref{tab:main_smpl} and~\ref{tab:smplh}) are at most $\pm0.02^{\circ}$ Geo, $\pm0.19$\,mm PA-MPJPE, $\pm0.002$ FID, and $\pm0.2$\,pp recall, one to two orders of magnitude below the Riemannian-Euclidean kinematic gaps. A Euclidean variant with per-step renormalization to the rotation manifold does not close the gap either, and instead destabilizes integration ($>44^{\circ}$ Geo for $N\geq20$).

\begin{table}[t]
    \centering
    \caption{\textbf{Effect of classifier-free guidance weight $w$ on whole-body (SMPL-H) generation} (all 52 joints, both main-paper variants, final checkpoints, seed 0). Retrieval rises monotonically with $w$ and kinematic error is near-minimal at $w{=}2$, the setting adopted in all main tables. The division of labour is unchanged across $w$: the Riemannian variant is better on PA-MPJPE and Geo at every $w$, the Euclidean variant leads FID throughout and retrieval for $w\,{\geq}\,2$ (Sec.~\ref{sec:exp_cfg}). Cell shading marks the best (red), second (orange), and third (yellow) value per column.}
    \label{tab:cfg}
    \small
    \setlength{\tabcolsep}{4pt}
    \resizebox{\linewidth}{!}{%
    \begin{tabular}{llccccc}
        \toprule
        Variant & $w$ & FID\,$\downarrow$ & R@1 T2P\,$\uparrow$ & R@1 P2T\,$\uparrow$
                & PA-MPJPE\,$\downarrow$ & Geo\,$\downarrow$ \\
                &     &      & (\%) & (\%) & (mm) & (deg) \\
        \midrule
        \multirow{3}{*}{Ours (Eucl.)}
          & 1.0 & 0.947 & 24.01 & 19.41 & 123.8 & 10.61 \\
          & 2.0 & \csecond{0.903} & \cthird{27.54} & \cthird{22.64} & \cthird{116.4} & 10.65 \\
          & 3.0 & \cbest{0.887} & \cbest{28.47} & \cbest{23.72} & 117.9 & 10.92 \\
        \midrule
        \multirow{3}{*}{Ours (Riem.)}
          & 1.0 & 0.960 & 24.46 & 19.22 & 120.7 & \csecond{10.39} \\
          & 2.0 & 0.942 & 27.42 & 22.29 & \cbest{111.6} & \cbest{10.31} \\
          & 3.0 & \cthird{0.932} & \csecond{27.74} & \csecond{22.86} & \csecond{112.6} & \cthird{10.52} \\
        \bottomrule
    \end{tabular}}
\end{table}

\subsubsection{Classifier-Free Guidance}
\label{sec:exp_cfg}

The guidance weight $w$ trades text alignment against rotational fidelity. We sweep $w$ at fixed $N=50$ (Table~\ref{tab:cfg}). Increasing $w$ improves both recall directions monotonically, while PA-MPJPE is minimised at $w=2$ for both variants and rises beyond it, as is geodesic error for the Riemannian model, whereas the Euclidean model's geodesic error is lowest at $w=1$. We therefore adopt $w=2$ in all main-table results, which retains near-minimal kinematic error while recovering most recall gains. The division of labour is unchanged across $w$: the Riemannian variant attains lower PA-MPJPE and geodesic error at every $w$, while the Euclidean variant leads FID throughout and retrieval for $w\geq2$, indicating that the manifold formulation's kinematic advantage is not specific to a particular guidance setting.

\begin{table}[t]
    \centering
    \caption{\textbf{Architecture and training-data ablations} (whole-body SMPL-H, seed 0). Each row varies one axis from the default model (part-tokenized, PoseScript only): the Joint rows use per-joint attention, the PS+MX++ rows add Motion-X++ supervision. Part tokenization wins the kinematic metrics, per-joint attention the remaining columns, and MX++ costs kinematics at both architectures (Sec.~\ref{sec:exp_arch}). Cell shading marks the best (red), second (orange) and third (yellow) value per column.}
    \label{tab:ablation_arch}
    \setlength{\tabcolsep}{4pt}
    \small
    \resizebox{\linewidth}{!}{%
    \begin{tabular}{llccccc}
        \toprule
        Token & Train Data & FID\,$\downarrow$ & R@1 T2P\,$\uparrow$ & R@1 P2T\,$\uparrow$ & PA-MPJPE\,$\downarrow$ & Geo\,$\downarrow$ \\
                     &               &      & (\%) & (\%) & (mm) & (deg) \\
        \midrule
        Joint & PS only & \cbest{0.926} & \cbest{28.18} & \cbest{22.82} & \cthird{114.1} & \cthird{10.60} \\
        Joint & PS+MX++ & \csecond{0.929} & \csecond{28.11} & \csecond{22.65} & 117.9 & 10.85 \\
        Part & PS only & \cthird{0.942} & 27.42 & 22.29 & \cbest{111.6} & \cbest{10.31} \\
        Part & PS+MX++ & 0.950 & \cthird{27.91} & \cthird{22.44} & \csecond{112.6} & \csecond{10.54} \\
        \bottomrule
    \end{tabular}}
\end{table}

\begin{table}[t]
    \centering
    \caption{\textbf{Text-conditioning ablation} (whole-body SMPL-H, Riemannian variant, seed 0). Swapping the encoder under pooled conditioning barely changes any metric, whereas all-token cross-attention lifts every one (Sec.~\ref{sec:exp_arch}). Cell shading marks the best (red), second (orange) and third (yellow) value per column.}
    \label{tab:cond}
    \setlength{\tabcolsep}{4pt}
    \small
    \resizebox{\linewidth}{!}{%
        \begin{tabular}{llccccc}
        \toprule
        Encoder & Cond. & FID\,$\downarrow$ & R@1 T2P\,$\uparrow$ & R@1 P2T\,$\uparrow$ & PA-MPJPE\,$\downarrow$ & Geo\,$\downarrow$ \\
                &              &      & (\%)   & (\%)   & (mm) & (deg) \\
        \midrule
        Qwen3-Emb-4B & pooled & \cthird{0.978} & \csecond{18.05} & \csecond{14.22} & \csecond{163.0} & \csecond{12.09} \\
        Flan-T5-Base & pooled & \csecond{0.974} & \cthird{17.48} & \cthird{14.04} & \cthird{164.4} & \cthird{12.17} \\
        Flan-T5-Base & cross & \cbest{0.942} & \cbest{27.42} & \cbest{22.29} & \cbest{111.6} & \cbest{10.31} \\
        \bottomrule
        \end{tabular}
    }
\end{table}

\subsubsection{Architecture, Conditioning, and Training Data}
\label{sec:exp_arch}

We ablate the remaining design choices (part tokenization, text conditioning, and Motion-X++ supervision) by varying one axis at a time from the default model (Tables~\ref{tab:ablation_arch} and~\ref{tab:cond}). Replacing the eight part tokens with per-joint tokenization at matched data yields a trade-off rather than a uniform loss: part tokenization wins both kinematic metrics (PA-MPJPE $111.6$ vs.\ $114.1$\,mm, Geo $10.31^\circ$ vs.\ $10.60^\circ$), while per-joint attention is slightly better on FID and retrieval at $6.5\times$ the sequence length. The batch-matched pair (the PS+MX++ rows) shows the same kinematic pattern ($112.6$ vs.\ $117.9$\,mm, $10.54^\circ$ vs.\ $10.85^\circ$). Part tokenization therefore improves the kinematic fidelity targeted by our objective while reducing the attention sequence length from $52$ to $8$ tokens. Replacing all-token cross-attention with a pooled text embedding substantially degrades alignment and reconstruction regardless of encoder capacity (Table~\ref{tab:cond}). Swapping the $4$B embedding encoder for the 250M Flan-T5-Base under pooled conditioning barely changes any metric, while enabling cross-attention at the same encoder lifts T2P recall from $17.5$ to $27.4\,\%$ and reduces PA-MPJPE by $53$\,mm. The conditioning pathway, rather than encoder capacity, is the bottleneck. Adding Motion-X++ at fixed architecture does not improve reconstruction: it costs PA-MPJPE and geodesic error at both architectures ($112.6$ vs.\ $111.6$\,mm and $10.54^\circ$ vs.\ $10.31^\circ$ at the part architecture) and improves only the retrieval directions there. PoseScript contains roughly 100k poses, small relative to the scale typical of flow-based generative models, and Motion-X++ supplies additional whole-body coverage. Its automatically generated captions differ stylistically from the PoseScript descriptions our retrieval metrics are built on, so we cap its contribution at 300k poses. The absence of a kinematic gain indicates that the primary benefit of PoseRieFlow arises from the model formulation rather than increased training data, and we keep PoseScript-only training as the default.

\section{Conclusion}
\label{sec:conclusion}

We presented \emph{PoseRieFlow}, a Riemannian flow matching model for text-conditioned whole-body pose generation. By formulating the flow on the product manifold $\SO^{52}$ with constant-velocity geodesic targets, training a compact Diffusion Transformer with AdaLN time conditioning and cross-attention to Flan-T5-Base features, we outperform far larger LLM-based baselines on kinematic fidelity while modelling the full SMPL-H articulation. Ablations show that the Riemannian formulation consistently improves rotational and joint-position fidelity over Euclidean flow matching at matched compute across inference budgets and guidance weights, that token-level cross-attention rather than encoder capacity drives text alignment, and that part tokenization improves kinematic fidelity while cutting attention cost relative to per-joint tokenization.

\paragraph{Limitations and future work.}
Our captions and evaluation depend on PoseScript-style descriptions and Motion-X++'s automatically generated whole-body pose descriptions. Even PoseScript's human-written captions are detailed, body-part-relative descriptions written to discriminate the pose from distractors, so performance on free-form colloquial requests is not characterised, and dataset biases in AMASS and Motion-X++ may propagate to generated poses. We do not explore learned or adaptive groupings, nor granularities between eight parts and per-joint tokenization. We do not model body shape, facial expression, or contact constraints. As with any generative pose system, misuse risks arise if outputs are linked to identity or surveillance applications. Natural next steps include extending the flow to body shape, integrating contact priors as gradient guidance, and combining PoseRieFlow with a multimodal LLM front-end~\cite{li2025unipose} for caption generality at compact compute.

\section*{Acknowledgements}
This work was supported by the European Regional Development Fund (ERDF) and the State of North Rhine-Westphalia as part of the operational program EFRE/JTF-Programm NRW 2021-2027. The project, titled “Gen-AIvatar”, was funded under the NEXT.IN.NRW competition with the grant agreement No. EFRE-20801085.

The work has also been funded by the Ministry of Culture and Science North Rhine-Westphalia under grant number PB22-063A (InVirtuo 4.0: Experimental Research in Virtual Environments), and by the state of North Rhine Westphalia as part of the Excellency Start-up Center.NRW (U-BO-GROW) under grant number 03ESCNW18B.

Additionally, it has been funded by the Federal Ministry of Education and Research under Grant No. 01IS22094A WEST-AI.

{
    \small
    \bibliographystyle{ieeenat_fullname}
    \bibliography{main}

@String{CVPR = "IEEE/CVF Conference on Computer Vision and Pattern Recognition (CVPR)"}

@String{ICCV = "IEEE International Conference on Computer Vision (ICCV)"}

@String{ECCV = "European Conference on Computer Vision (ECCV)"}

@String{AAAI = "AAAI Conference on Artificial Intelligence"}

@String{ICML = "International Conference on Machine Learning (ICML)"}

@String{ICLR = "International Conference on Learning Representations (ICLR)"}

@String{IROS = "IEEE/RSJ International Conference on Intelligent Robots and Systems (IROS)"}

@String{SIGGRAPH = "Annual Conference on Computer Graphics and Interactive Techniques (SIGGRAPH)"}

@String{CHI = "ACM Conference on Human Factors in Computing Systems (CHI)"}

@String{THREE_DV = "International Conference on 3D Vision (3DV)"}

@String{BMVC = "British Machine Vision Conference (BMVC)"}

@inproceedings{
tevet2023human,
title={Human Motion Diffusion Model},
author={Guy Tevet and Sigal Raab and Brian Gordon and Yoni Shafir and Daniel Cohen-or and Amit Haim Bermano},
booktitle=ICLR,
year={2023},
}

@article{gao2025motionflux,
  title={MotionFlux: Efficient Text-Guided Motion Generation through Rectified Flow Matching and Preference Alignment},
  author={Gao, Zhiting and Song, Dan and Jiang, Diqiong and Xue, Chao and Liu, An-An},
  journal={arXiv preprint arXiv:2508.19527},
  year={2025}
}

@article{hymotion2025,
  title={HY-Motion 1.0: Scaling Flow Matching Models for Text-To-Motion Generation},
  author={Tencent Hunyuan 3D Digital Human Team},
  journal={arXiv preprint arXiv:2512.23464},
  year={2025}
}

@ARTICLE{delmas2024posescript,
  author={Delmas, Ginger and Weinzaepfel, Philippe and Lucas, Thomas and Moreno-Noguer, Francesc and Rogez, Grégory},
  journal={IEEE Transactions on Pattern Analysis and Machine Intelligence}, 
  title={PoseScript: Linking 3D Human Poses and Natural Language}, 
  year={2025},
  volume={47},
  number={7},
  pages={5146-5159},
  doi={10.1109/TPAMI.2024.3407570}}

@inproceedings{delmas2023posefix,
  title={PoseFix: Correcting 3D Human Poses with Natural Language},
  author={Delmas, Ginger and Weinzaepfel, Philippe and Moreno-Noguer, Francesc and Rogez, Gr{\'e}gory},
  booktitle=ICCV,
  year={2023}
}

@inproceedings{delmas2024poseembroider,
  title={PoseEmbroider: Towards a 3D, Visual, Semantic-aware Human Pose Representation},
  author={Delmas, Ginger and Weinzaepfel, Philippe and Moreno-Noguer, Francesc and Rogez, Gr{\'e}gory},
  booktitle=ECCV,
  year={2024}
}

@inproceedings{feng2024chatpose,
  title = {{ChatPose}: Chatting about 3D Human Pose},
  booktitle = {IEEE/CVF Conf.~on Computer Vision and Pattern Recognition (CVPR)},
  pages = {2093--2103},
  address = {Piscataway, NJ},
  month = sep,
  year = {2024},
  author = {Feng, Yao and Lin, Jing and Dwivedi, Sai Kumar and Sun, Yu and Patel, Priyanka and Black, Michael J.},
  doi = {10.1109/CVPR52733.2024.00204},
  url = {https://yfeng95.github.io/ChatPose/},
  month_numeric = {9}
}

@inproceedings{posellava2025,
  title={PoseLLaVA: Pose Centric Multimodal LLM for Fine-Grained 3D Pose Manipulation},
  author={Feng, Dong and Guo, Ping and Peng, Encheng and Zhu, Mingmin and Yu, Wenhao and Wang, Peng},
  booktitle=AAAI,
  volume={39},
  pages={2951--2959},
  year={2025}
}

@InProceedings{li2025unipose,
    author    = {Li, Yiheng and Hou, Ruibing and Chang, Hong and Shan, Shiguang and Chen, Xilin},
    title     = {UniPose: A Unified Multimodal Framework for Human Pose Comprehension, Generation and Editing},
    booktitle = CVPR,
    month     = {June},
    year      = {2025},
    pages     = {27805-27815}
}

@InProceedings{cha2025cotpose,
    author    = {Cha, Junuk and Kim, Jihyeon},
    title     = {CoT-Pose: Chain-of-Thought Reasoning for 3D Pose Generation from Abstract Prompts},
    booktitle = {Proceedings of the IEEE/CVF International Conference on Computer Vision (ICCV) Workshops},
    month     = {October},
    year      = {2025},
    pages     = {4360-4368}
}

@inproceedings{he2024nrdf,
    title = {NRDF: Neural Riemannian Distance Fields for Learning Articulated Pose Priors},
    author = {He, Yannan and Tiwari, Garvita and Birdal, Tolga and Lenssen, Jan Eric and Pons-Moll, Gerard},
    booktitle = CVPR,
    year = {2024},
}

@InProceedings{yu2025geometric,
    author    = {Yu, Zhengdi and Foti, Simone and Zhang, Linguang and Zhao, Amy and Keskin, Cem and Zafeiriou, Stefanos and Birdal, Tolga},
    title     = {Geometric Neural Distance Fields for Learning Human Motion Priors},
    booktitle = {Proceedings of the IEEE/CVF Conference on Computer Vision and Pattern Recognition (CVPR)},
    month     = {June},
    year      = {2026},
    pages     = {2232-2242}
}

@article{miao2026riemannian,
  title={Riemannian Motion Generation: A Unified Framework for Human Motion Representation and Generation via Riemannian Flow Matching},
  author={Miao, Fangran and Huang, Jian and Li, Ting},
  journal={arXiv preprint arXiv:2603.15016},
  year={2026}
}

@inproceedings{
lipman2023flow,
title={Flow Matching for Generative Modeling},
author={Yaron Lipman and Ricky T. Q. Chen and Heli Ben-Hamu and Maximilian Nickel and Matthew Le},
booktitle=ICLR,
year={2023},
}

@inproceedings{
chen2024flow,
title={Flow Matching on General Geometries},
author={Ricky T. Q. Chen and Yaron Lipman},
booktitle=ICLR,
year={2024},
}

@inproceedings{bose2024se3,
  title={SE (3)-Stochastic Flow Matching for Protein Backbone Generation},
  author={Bose, Joey and Akhound-Sadegh, Tara and Huguet, Guillaume and FATRAS, Kilian and Rector-Brooks, Jarrid and Liu, Cheng-Hao and Nica, Andrei Cristian and Korablyov, Maksym and Bronstein, Michael M and Tong, Alexander},
  booktitle=ICLR,
    year={2024}
}

@inproceedings{yim2023se3,
author = {Yim, Jason and Trippe, Brian L. and De Bortoli, Valentin and Mathieu, Emile and Doucet, Arnaud and Barzilay, Regina and Jaakkola, Tommi},
title = {SE(3) diffusion model with application to protein backbone generation},
year = {2023},
booktitle = ICML,
}

@inproceedings{braun2024riemannianfm,
  title={Riemannian flow matching policy for robot motion learning},
  author={Braun, Max and Jaquier, No{\'e}mie and Rozo, Leonel and Asfour, Tamim},
  booktitle=IROS,
  year={2024},
}

@article{hu2023motion,
  title={Motion flow matching for human motion synthesis and editing},
  author={Hu, Vincent Tao and Yin, Wenzhe and Ma, Pingchuan and Chen, Yunlu and Fernando, Basura and Asano, Yuki M and Gavves, Efstratios and Mettes, Pascal and Ommer, Bjorn and Snoek, Cees GM},
  journal={arXiv preprint arXiv:2312.08895},
  year={2023}
}

@inproceedings{narasimhaswamy2024handiffuser,
  title={Handiffuser: Text-to-image generation with realistic hand appearances},
  author={Narasimhaswamy, Supreeth and Bhattacharya, Uttaran and Chen, Xiang and Dasgupta, Ishita and Mitra, Saayan and Hoai, Minh},
  booktitle=CVPR,
  year={2024}
}

@article{zhang2025motion,
  title={Motion-X++: A Large-Scale Multimodal 3D Whole-body Human Motion Dataset},
  author={Zhang, Yuhong and Lin, Jing and Zeng, Ailing and Wu, Guanlin and Lu, Shunlin and Fu, Yurong and Cai, Yuanhao and Zhang, Ruimao and Wang, Haoqian and Zhang, Lei},
  journal={arXiv preprint arXiv:2501.05098},
  year={2025}
}

@inproceedings{peebles2023dit,
  title={Scalable diffusion models with transformers},
  author={Peebles, William and Xie, Saining},
  booktitle=ICCV,
  year={2023}
}

@inproceedings{esser2024sd3,
  title={Scaling Rectified Flow Transformers for High-Resolution Image Synthesis},
  author={Esser, Patrick and others},
  booktitle=ICML,
  year={2024}
}

@inproceedings{zhou20196d,
  title={On the Continuity of Rotation Representations in Neural Networks},
  author={Zhou, Yi and Barnes, Connelly and Lu, Jingwan and Yang, Jimei and Li, Hao},
  booktitle=CVPR,
  year={2019}
}

@article{chung2024t5,
author = {Chung, Hyung Won and Hou, Le and Longpre, Shayne and Zoph, Barret and Tai, Yi and Fedus, William and Li, Yunxuan and Wang, Xuezhi and Dehghani, Mostafa and Brahma, Siddhartha and Webson, Albert and Gu, Shixiang Shane and Dai, Zhuyun and Suzgun, Mirac and Chen, Xinyun and Chowdhery, Aakanksha and Castro-Ros, Alex and Pellat, Marie and Robinson, Kevin and Valter, Dasha and Narang, Sharan and Mishra, Gaurav and Yu, Adams and Zhao, Vincent and Huang, Yanping and Dai, Andrew and Yu, Hongkun and Petrov, Slav and Chi, Ed H. and Dean, Jeff and Devlin, Jacob and Roberts, Adam and Zhou, Denny and Le, Quoc V. and Wei, Jason},
title = {Scaling instruction-finetuned language models},
year = {2024},
issue_date = {January 2024},
publisher = {JMLR.org},
volume = {25},
number = {1},
issn = {1532-4435},
journal = {J. Mach. Learn. Res.},
month = jan,
articleno = {70},
numpages = {53}
}

@InProceedings{guo2022humanml3d,
    author    = {Guo, Chuan and Zou, Shihao and Zuo, Xinxin and Wang, Sen and Ji, Wei and Li, Xingyu and Cheng, Li},
    title     = {Generating Diverse and Natural 3D Human Motions From Text},
    booktitle = {Proceedings of the IEEE/CVF Conference on Computer Vision and Pattern Recognition (CVPR)},
    month     = {June},
    year      = {2022},
    pages     = {5152-5161}
}

@inproceedings{leach2022so3,
  title={Denoising diffusion probabilistic models on {SO}(3) for rotational alignment},
  author={Leach, Adam and Schmon, Sebastian M and Degiacomi, Matteo T and Willcocks, Chris G},
  booktitle={ICLR Workshop on Geometrical and Topological Representation Learning},
  year={2022}}

@inproceedings{liu2023rectified,
  title={Flow straight and fast: Learning to generate and transfer data with rectified flow},
  author={Liu, Xingchao and Gong, Chengyue and Liu, Qiang},
  booktitle={International Conference on Learning Representations (ICLR)},
  year={2023}}

@inproceedings{zhang2024both2hands,
  title={{BOTH2Hands}: Inferring 3D hands from both text prompts and body dynamics},
  author={Zhang, Wenqian and Huang, Molin and Zhou, Yuxuan and Zhang, Juze and Yu, Jingyi and Wang, Jingya and Xu, Lan},
  booktitle={IEEE/CVF Conference on Computer Vision and Pattern Recognition (CVPR)},
  year={2024}}

@inproceedings{bian2025motioncraft,
  title={{MotionCraft}: Crafting whole-body motion with plug-and-play multimodal controls},
  author={Bian, Yuxuan and Zeng, Ailing and Ju, Xuan and Liu, Xian and Zhang, Zhaoyang and Liu, Wei and Xu, Qiang},
  booktitle={AAAI Conference on Artificial Intelligence},
  year={2025}}

@inproceedings{wang2023iip,
  title     = {{IIP-Transformer}: Intra-Inter-Part Transformer for Skeleton-Based Action Recognition},
  author    = {Wang, Qingtian and Shi, Shuze and He, Jiabin and Peng, Jianlin and Liu, Tingxi and Weng, Renliang},
  booktitle = {IEEE International Conference on Big Data (BigData)},
  pages     = {936--945},
  year      = {2023},
  doi       = {10.1109/BigData59044.2023.10386970}
}

@inproceedings{dou2023tore,
  title     = {{TORE}: Token Reduction for Efficient Human Mesh Recovery with Transformer},
  author    = {Dou, Zhiyang and Wu, Qingxuan and Lin, Cheng and Cao, Zeyu and Wu, Qiangqiang and Wan, Weilin and Komura, Taku and Wang, Wenping},
  booktitle = ICCV,
  pages     = {15143--15155},
  year      = {2023}
}

@inproceedings{wang2024gtpt,
  title     = {{GTPT}: Group-based Token Pruning Transformer for Efficient Human Pose Estimation},
  author    = {Wang, Haonan and Liu, Jie and Tang, Jie and Wu, Gangshan and Xu, Bo and Chou, Yanbing and Wang, Yong},
  booktitle = ECCV,
  year      = {2024}
}

@inproceedings{aksan2021sttransformer,
  title     = {A Spatio-temporal Transformer for {3D} Human Motion Prediction},
  author    = {Aksan, Emre and Kaufmann, Manuel and Cao, Peng and Hilliges, Otmar},
  booktitle = THREE_DV,
  year      = {2021}
}

@inproceedings{li2021tokenpose,
  title     = {{TokenPose}: Learning Keypoint Tokens for Human Pose Estimation},
  author    = {Li, Yanjie and Zhang, Shoukui and Wang, Zhicheng and Yang, Sen and Yang, Wankou and Xia, Shu-Tao and Zhou, Erjin},
  booktitle = ICCV,
  pages     = {11313--11322},
  year      = {2021}
}

@inproceedings{zhang2023finemogen,
author = {Zhang, Mingyuan and Li, Huirong and Cai, Zhongang and Ren, Jiawei and Yang, Lei and Liu, Ziwei},
title = {FineMoGen: fine-grained spatio-temporal motion generation and editing},
year = {2023},
publisher = {Curran Associates Inc.},
address = {Red Hook, NY, USA},
booktitle = {Proceedings of the 37th International Conference on Neural Information Processing Systems},
articleno = {616},
numpages = {12},
location = {New Orleans, LA, USA},
series = {NIPS '23}
}

@inproceedings{zou2024parco,
  title     = {{ParCo}: Part-Coordinating Text-to-Motion Synthesis},
  author    = {Zou, Qiran and Yuan, Shangyuan and Du, Shian and Wang, Yu and Liu, Chang and Xu, Yi and Chen, Jie and Ji, Xiangyang},
  booktitle = ECCV,
  year      = {2024}
}

@inproceedings{zhang2024lmm,
  title     = {Large Motion Model for Unified Multi-Modal Motion Generation},
  author    = {Zhang, Mingyuan and Jin, Daisheng and Gu, Chenyang and Hong, Fangzhou and Cai, Zhongang and Huang, Jingfang and Zhang, Chongzhi and Guo, Xinying and Yang, Lei and He, Ying and Liu, Ziwei},
  booktitle = ECCV,
  year      = {2024}
}

@inproceedings{siyao2022bailando,
  title     = {{Bailando}: {3D} Dance Generation by Actor-Critic {GPT} with Choreographic Memory},
  author    = {Siyao, Li and Yu, Weijiang and Gu, Tianpei and Lin, Chunze and Wang, Quan and Qian, Chen and Loy, Chen Change and Liu, Ziwei},
  booktitle = CVPR,
  pages     = {11050--11059},
  year      = {2022}
}

@inproceedings{zhong2023attt2m,
  title     = {{AttT2M}: Text-Driven Human Motion Generation with Multi-Perspective Attention Mechanism},
  author    = {Zhong, Chongyang and Hu, Lei and Zhang, Zihao and Xia, Shihong},
  booktitle = ICCV,
  pages     = {509--519},
  year      = {2023}
}

@inproceedings{yi2023talkshow,
  title     = {Generating Holistic {3D} Human Motion from Speech},
  author    = {Yi, Hongwei and Liang, Hualin and Liu, Yifei and Cao, Qiong and Wen, Yandong and Bolkart, Timo and Tao, Dacheng and Black, Michael J.},
  booktitle = CVPR,
  pages     = {469--480},
  year      = {2023}
}

@inproceedings{liu2024emage,
  title     = {{EMAGE}: Towards Unified Holistic Co-Speech Gesture Generation via Expressive Masked Audio Gesture Modeling},
  author    = {Liu, Haiyang and Zhu, Zihao and Becherini, Giorgio and Peng, Yichen and Su, Mingyang and Zhou, You and Zhe, Xuefei and Iwamoto, Naoya and Zheng, Bo and Black, Michael J.},
  booktitle = CVPR,
  pages     = {1144--1154},
  year      = {2024}
}

@inproceedings{lu2024humantomato,
  title     = {{HumanTOMATO}: Text-aligned Whole-body Motion Generation},
  author    = {Lu, Shunlin and Chen, Ling-Hao and Zeng, Ailing and Lin, Jing and Zhang, Ruimao and Zhang, Lei and Shum, Heung-Yeung},
  booktitle = ICML,
  year      = {2024}
}

@article{zhang2026samor,
  title   = {{SAMoR}: Motion Modelling for Articulated Objects of Any Skeleton and Topology},
  author  = {Zhang, Yuhao and Pons-Moll, Gerard and Birdal, Tolga},
  journal = {arXiv preprint arXiv:2607.02148},
  year    = {2026}
}

@inproceedings{zeng2022tcformer,
  title     = {Not All Tokens Are Equal: Human-centric Visual Analysis via Token Clustering Transformer},
  author    = {Zeng, Wang and Jin, Sheng and Liu, Wentao and Qian, Chen and Luo, Ping and Ouyang, Wanli and Wang, Xiaogang},
  booktitle = CVPR,
  pages     = {11101--11111},
  year      = {2022}
}

@inproceedings{du2015hierarchical,
  title     = {Hierarchical Recurrent Neural Network for Skeleton Based Action Recognition},
  author    = {Du, Yong and Wang, Wei and Wang, Liang},
  booktitle = CVPR,
  pages     = {1110--1118},
  year      = {2015}
}

@inproceedings{jain2016srnn,
  title     = {{Structural-RNN}: Deep Learning on Spatio-Temporal Graphs},
  author    = {Jain, Ashesh and Zamir, Amir R. and Savarese, Silvio and Saxena, Ashutosh},
  booktitle = CVPR,
  pages     = {5308--5317},
  year      = {2016}
}

@inproceedings{thakkar2018pbgcn,
  title     = {Part-based Graph Convolutional Network for Action Recognition},
  author    = {Thakkar, Kalpit C. and Narayanan, P. J.},
  booktitle = BMVC,
  year      = {2018}
}

@article{romero2017mano,
  title = {Embodied Hands: Modeling and Capturing Hands and Bodies Together},
  author = {Romero, Javier and Tzionas, Dimitrios and Black, Michael J.},
  journal = {ACM Transactions on Graphics, (Proc. SIGGRAPH Asia)},
  volume = {36},
  number = {6},
  series = {245:1--245:17},
  month = nov,
  year = {2017},
  month_numeric = {11}
}

@article{looper2015smpl,
author = {Loper, Matthew and Mahmood, Naureen and Romero, Javier and Pons-Moll, Gerard and Black, Michael J.},
title = {SMPL: a skinned multi-person linear model},
year = {2015},
issue_date = {November 2015},
publisher = {Association for Computing Machinery},
address = {New York, NY, USA},
volume = {34},
number = {6},
issn = {0730-0301},
url = {https://doi.org/10.1145/2816795.2818013},
doi = {10.1145/2816795.2818013},
journal = {ACM Trans. Graph.},
month = nov,
articleno = {248},
numpages = {16}
}

@inproceedings{pavlakos2019smplx,
  title = {Expressive Body Capture: {3D} Hands, Face, and Body from a Single Image},
  author = {Pavlakos, Georgios and Choutas, Vasileios and Ghorbani, Nima and Bolkart, Timo and Osman, Ahmed A. A. and Tzionas, Dimitrios and Black, Michael J.},
  booktitle = {Proceedings IEEE Conf. on Computer Vision and Pattern Recognition (CVPR)},
  pages     = {10975--10985},
  year = {2019}
}

@conference{AMASS,
  title           = {{AMASS}: Archive of Motion Capture as Surface Shapes},
  author          = {Mahmood, Naureen and Ghorbani, Nima and Troje, Nikolaus F. and Pons-Moll, Gerard and Black, Michael J.},
  booktitle       = {International Conference on Computer Vision},
  pages           = {5442--5451},
  month           = oct,
  year            = {2019},
  month_numeric   = {10}
}

@inproceedings{yu2024motionpatches,
  author    = {Qing Yu and Mikihiro Tanaka and Kent Fujiwara},
  title     = {Exploring Vision Transformers for {3D} Human Motion-Language Models with Motion Patches},
  booktitle = {IEEE/CVF Conference on Computer Vision and Pattern Recognition (CVPR)},
  pages     = {937--946},
  year      = {2024}
}

@inproceedings{yu2025remogpt,
  author    = {Qing Yu and Mikihiro Tanaka and Kent Fujiwara},
  title     = {{ReMoGPT}: Part-Level Retrieval-Augmented Motion-Language Models},
  booktitle = {AAAI Conference on Artificial Intelligence},
  pages     = {9635--9643},
  year      = {2025}
}
}

\end{document}